\pdfoutput=1
\documentclass{article}

 \PassOptionsToPackage{numbers, sort&compress}{natbib}

\usepackage[final]{neurips_data_2021}

\usepackage[utf8]{inputenc} % allow utf-8 input
\usepackage[T1]{fontenc}    % use 8-bit T1 fonts
\usepackage[hidelinks]{hyperref}
\usepackage{url}            % simple URL typesetting
\usepackage{booktabs}       % professional-quality tables
\usepackage{amsfonts}       % blackboard math symbols
\usepackage{nicefrac}       % compact symbols for 1/2, etc.
\usepackage{microtype}      % microtypography
\usepackage{xcolor}         % colors

\usepackage{graphicx}
\usepackage{xcolor}
\usepackage{bm}
\usepackage{amssymb,amsmath,amsthm}
\usepackage{mathtools}
\usepackage{relsize}
\usepackage{multirow}
\usepackage{subcaption}
\usepackage{verbatim}
\usepackage{float}
\usepackage{enumitem}
\newcommand*{\todo}[2][]{\textcolor{red}{[\textbf{\ifthenelse{\equal{#1}{}}{TODO}{TODO(#1)}}: #2]}}

\newtheorem{definition}{Definition}

\DeclareMathOperator*{\argmax}{arg\,max}

\newcommand{\smalleq}{{\scriptstyle =}}

\newcommand{\smallin}{{\scriptstyle \in}}

\newcommand{\joint}{\bm{Q}_{\tilde{y}, y^*}}

\newcommand{\estjointlong}{\hat{\bm{Q}}_{\tilde{y} = i, y^* = j}}

\newcommand{\probmatrix}{\bm{\hat{P}}_{k,i}}

\newcommand{\predprobshortj}{\hat{p}_{\bm{x}, \tilde{y} \smalleq j}}

\newcommand{\perfprobshortj}{p^*_{\bm{x}, \tilde{y} \smalleq j}}
\newcommand{\perfprobshortk}{p^*_{\bm{x}, \tilde{y} \smalleq k}}

\newcommand{\errorxj}{\epsilon_{\bm{x}, \tilde{y} \smalleq j}}
\newcommand{\errorxk}{\epsilon_{\bm{x}, \tilde{y} \smalleq k}}
\newcommand{\cj}{\bm{C}_{\tilde{y}, y^*}}

\newcommand{\estpartition}{\hat{\bm{X}}_{ \tilde{y} \smalleq i,y^* \smalleq j}}
\newcommand{\partition}{\bm{X}_{ \tilde{y} \smalleq i,y^* \smalleq j}}

\newcommand{\beginsupplement}{ % use to mark beginning of supplementary section.
\setcounter{section}{0}
\renewcommand{\thesection}{S\arabic{section}} %
\renewcommand{\thesubsection}{\thesection.\arabic{subsection}}
\setcounter{table}{0}
\renewcommand{\thetable}{S\arabic{table}} %
\setcounter{figure}{0}
\renewcommand{\thefigure}{S\arabic{figure}} %
}

\definecolor{darkblue}{HTML}{091c70}
\definecolor{highlightcolor}{HTML}{ffdada}
\usepackage{hyperref}
\hypersetup{
    colorlinks=True,
    linkcolor=darkblue,
    filecolor=darkblue,      
    urlcolor=darkblue,
    citecolor=darkblue,
}

\newcommand{\papertitle}{Pervasive Label Errors in Test Sets\\ Destabilize Machine Learning Benchmarks}

\title{\papertitle}

\author{%
  Curtis G.~Northcutt\thanks{Correspondence to: \texttt{curtis@cleanlab.ai} or \texttt{cgn@csail.mit.edu}.} \\
  ChipBrain, MIT, Cleanlab \\
  \And
  Anish Athalye \\
  MIT, Cleanlab \\
  \And
  Jonas Mueller \\
  AWS \\
}

\begin{document}

\maketitle

\setcounter{footnote}{0}

\begin{abstract}
  We identify label errors in the \emph{test} sets of 10 of the most commonly-used computer vision, natural language, and audio datasets, and subsequently study the potential for these label errors to affect benchmark results. 
Errors in test sets are numerous and widespread: we estimate an average of at least 3.3\% errors across the 10 datasets, where for example label errors comprise at least 6\% of the ImageNet validation set. 
Putative label errors are identified using confident learning algorithms and then human-validated via crowdsourcing (51\% of the algorithmically-flagged candidates are indeed erroneously labeled, on average across the datasets). 
Traditionally, machine learning practitioners choose
which model to deploy based on test accuracy 
--- our findings advise caution here, proposing that judging models over correctly labeled test sets may be more useful, especially for noisy real-world datasets. 
Surprisingly, we find that lower capacity models may be practically more useful than higher capacity models in real-world datasets with high proportions of erroneously labeled data.
For example, on ImageNet with corrected labels: ResNet-18 outperforms ResNet-50 if the prevalence of originally mislabeled test examples increases by just 6\%. On CIFAR-10 with corrected labels:  VGG-11 outperforms VGG-19 if the prevalence of originally mislabeled test examples increases by just 5\%. 
Test set errors across the 10 datasets can be viewed at \url{https://labelerrors.com} and all label errors can be reproduced by \url{https://github.com/cleanlab/label-errors}.

\end{abstract}

\section{Introduction}

Large labeled datasets have been critical to the success of supervised
machine learning across the board in domains such as image classification,
sentiment analysis, and audio classification.
Yet, the processes used to construct datasets
often involve some degree of automatic labeling or crowd-sourcing, techniques
which are inherently error-prone \citep{chi2021data}. Even with controls for error
correction~\citep{pmlr-robust-active-label-correction-2018, zhang2017improving}, errors can slip through.
Prior work has considered the consequences of noisy labels, usually in the
context of \emph{learning} with noisy labels, and usually focused on noise in the \emph{train} set.
Some past research has concluded that label noise is not a major concern, because of
techniques to learn with noisy labels~\citep{patrini2017making, NIPS2013_5073},
and also because deep learning is believed to be naturally robust to label noise~\citep{rolnick2017deep, gupta_unreasonable_effectiveness_of_data_2017, huang2019understanding_generalization, fair_laurens_van_der_maaten_limits_weak_supervision_2018}.

However, label errors in \emph{test} sets are less-studied and have a different set of potential
consequences.
Whereas \emph{train} set labels in a small number of machine learning datasets, e.g. in the ImageNet dataset, are well-known to contain errors \citep{northcutt2021confidentlearning, imageneterror2020shankar, hooker2019selective}, labeled data in \emph{test} sets is often considered ``correct'' as long as it is drawn from the same distribution as the train set. This is a fallacy: machine learning \emph{test} sets can, and do, contain errors, and these errors can destabilize ML benchmarks.

Researchers rely on benchmark test datasets to evaluate and
measure progress in the state-of-the-art and to validate theoretical findings.
If label errors occurred profusely, they could potentially undermine the
framework by which we measure progress in machine learning. Practitioners rely
on their own real-world datasets which are often more noisy than
carefully-curated benchmark datasets. Label errors in these test sets could
potentially lead practitioners to incorrect conclusions about which models
actually perform best in the real world.

\begin{figure*}
    \centering
    \includegraphics[width=\linewidth]{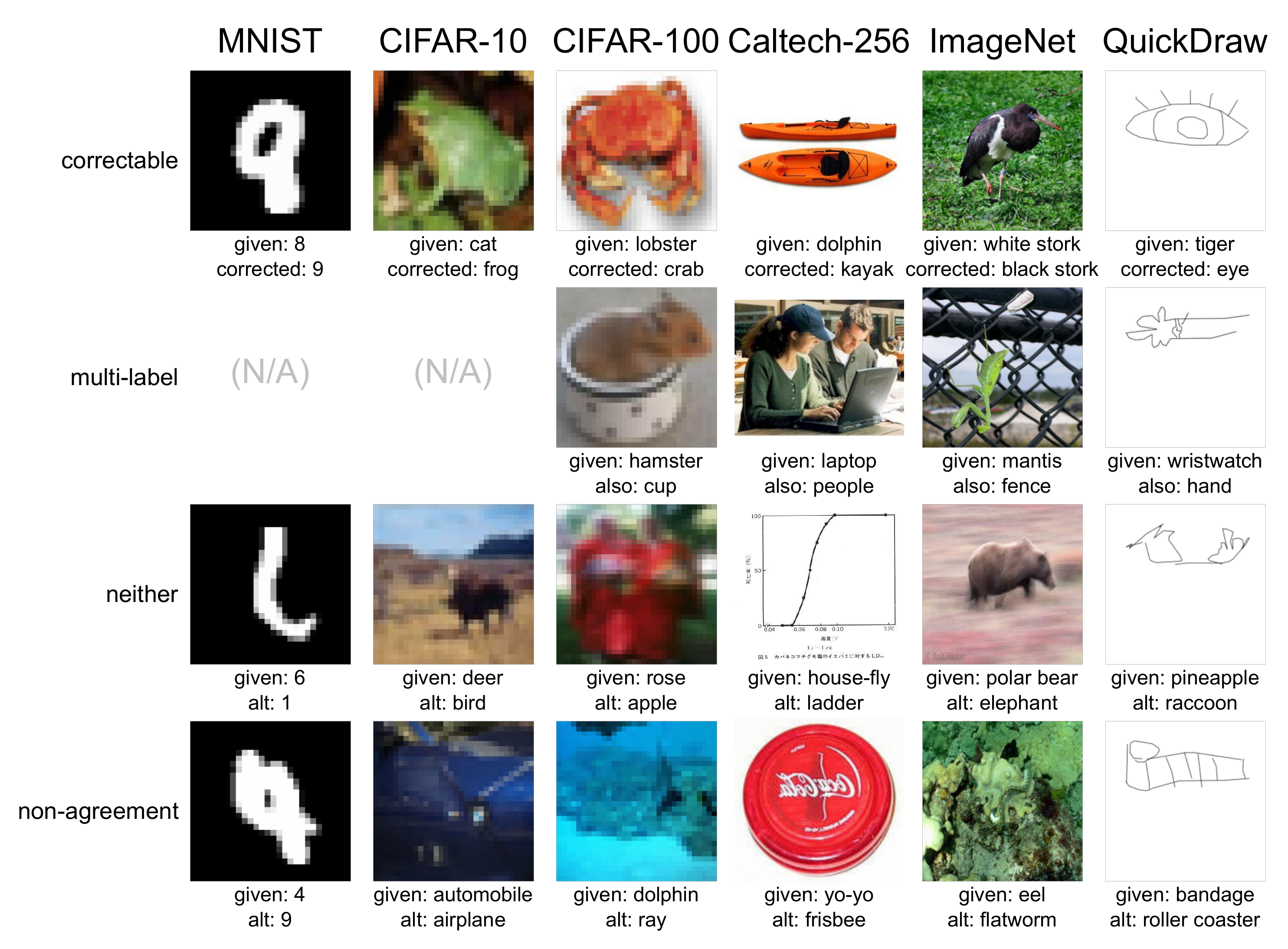}
    \caption{An example label error from each category (Section~\ref{validating}) for image datasets. The figure shows \underline{given} labels, human-validated \underline{corrected} labels, \underline{also} the second label for multi-class data points, and CL-guessed \underline{alt}ernatives. A gallery of label errors across all 10 datasets, including text and audio datasets, is available at \url{https://labelerrors.com}.}
    \label{fig:image-val}
\end{figure*}

We present the first study that systematically characterizes label 
errors across 10 datasets commonly used for benchmarking models in computer vision, natural
language processing, and audio processing. Unlike prior work on noisy labels,
we do not experiment with synthetic noise but with naturally-occurring errors. 
Rather than exploring a novel methodology for dealing with label errors, which has been extensively studied in the literature \citep{survery_deep_learning_noisy_labels}, this paper aims to characterize the prevalence of label errors in the test data of popular benchmarks used to measure ML progress and subsequently analyze practical consequences of these errors, and in particular, their effects on model selection. 
Using \emph{confident learning} \citep{northcutt2021confidentlearning}, we
algorithmically identify putative label errors in test sets at scale, and we validate these label errors through human evaluation, estimating a lower-bound of 3.3\% errors on average across the 10 datasets. We identify,
for example, 2916 (6\%) errors in the ImageNet validation set (which is
\emph{commonly used as a test set}), and estimate over 5 million (10\%) errors
in QuickDraw. Figure~\ref{fig:image-val} shows examples of validated label
errors for the image datasets in our study.

We use ImageNet and CIFAR-10 as case studies to understand the consequences of
test set label errors on benchmark stability. While there are numerous % (2916, 6\%) 
erroneous labels in 
these benchmarks' test data, we find that relative rankings of models
in benchmarks are unaffected after removing or correcting these label errors. 
However, we find that these benchmark results are \emph{unstable}: 
higher-capacity models (like NASNet) undesirably reflect the distribution of
systematic label errors in their predictions to a greater degree than
models with fewer parameters (like ResNet-18), and this effect \emph{increases}
with the prevalence of mislabeled test data. This is not traditional overfitting.
Larger models are able to generalize better to the given noisy labels in the
test data, but this is problematic because these models produce \emph{worse} predictions
than their lower-capacity counterparts when evaluated on the corrected labels
for originally-mislabeled test examples. 

In real-world settings with high proportions
of erroneously labeled data, lower capacity models may thus be practically more
useful than their higher capacity counterparts. For example, it may appear NASNet is superior to ResNet-18 based on the test accuracy over originally given labels, but NASNet is in fact worse than ResNet-18 based on the test accuracy over corrected labels. Since the latter form of accuracy is what matters in practice, ResNet-18 should actually be deployed instead of NASNet here --- but this is unknowable without correcting the test data labels.

To evaluate how benchmarks of popular pre-trained models change, we incrementally increase the noise prevalence by controlling for the proportion of correctable (but originally mislabeled) data within the test dataset. This procedure allows us to determine, for a particular dataset, at what noise prevalence benchmark rankings change. For example, on ImageNet with corrected labels: ResNet-18 outperforms ResNet-50 if the prevalence of originally mislabeled test examples increases by just 6\%.

In summary, our contributions include:

\begin{enumerate}
    \item The discovery of pervasive label errors in test sets of 10 standard ML benchmarks
    \item Open-sourced resources to clean and correct each test set, in which a large fraction of the label errors have been corrected by humans
    \item An analysis of the implications of test set label errors on benchmarks, and the finding that higher-capacity models perform better on the subset of incorrectly-labeled test data in terms of their accuracy on the original labels (i.e.,\ what one traditionally measures), but perform worse on this subset in terms of their accuracy on corrected labels (i.e.,\ what one cares about in practice, but cannot measure without the corrected test data we provide)
    \item The discovery that merely slight increases in the test label error prevalence would cause model selection to favor the wrong model based on standard test accuracy benchmarks
\end{enumerate}

Our findings imply ML practitioners might benefit from correcting test set labels to benchmark how their models will perform in real-world deployment, and by using simpler/smaller models in applications where labels for their datasets tend to be noisier than the labels in gold-standard benchmark datasets. One way to ascertain whether a dataset is noisy enough to suffer from this effect is to correct at least the test set labels, e.g.\ using our straightforward approach.

\section{Background and related work}

Experiments in learning with noisy labels \citep{DBLP:conf/icml/PatriniNNC16, rooyen_menon_unhinged_nips15, NIPS2013_5073, 7837934_multiclass_learning_with_noise_using_dropout, Sukhbaatar_fergus_iclr_2015} suffer a double-edged sword: either synthetic noise must be added to clean training data to measure performance on a clean test set (at the expense of modeling \emph{actual} real-world label noise \citep{pmlr-v119-jiang20c-synthetic-noise}), or a naturally noisy dataset is used and accuracy is measured on a noisy test set. In the noisy WebVision dataset \citep{li2017webvision}, accuracy on the ImageNet validation data is often reported as a ``clean'' test set, but several studies  \citep{recht2019imagenet, northcutt2021confidentlearning, tsipras2020imagenet, hooker2019selective} have shown the existence of label issues in ImageNet. Unlike these works, we focus exclusively on existence and implications of label errors in the test set, and we extend our analysis to many types of datasets. Although extensive prior work deals with label errors in the \emph{training} set \citep{Frenay2014, survery_deep_learning_noisy_labels}, much less work has been done to understand the implications of label errors in the \emph{test set}. 

Crowd-sourced curation of labels via multiple human workers \citep{zhang2017improving, dawid1979maximum, ratner_chris_re_large_training_sets_NIPS2016_6523} is a common method for validating/correcting label issues in datasets, but it can be exorbitantly expensive for large datasets. To circumvent this issue, we only validate subsets of datasets by first estimating which examples are most likely to be mislabeled. To achieve this, we lean on a number of contributions in uncertainty quantification for finding label errors based on prediction/label agreement via confusion matrices \citep{neurips2019novelinformationtheory, hendrycks2018using, chen2019confusion, icml_lipton_label_shift_confusion_matrix}; however, these approaches lack either robustness to class imbalance or theoretical support for realistic settings with \emph{asymmetric, non-uniform noise} (for instance, an image of a dog might be more likely to be mislabeled a coyote than a car). For robustness to class imbalance and theoretical support for exact uncertainty quantification, we use a model-agnostic framework, confident learning (CL) \citep{northcutt2021confidentlearning}, to estimate which labels are erroneous across diverse datasets.
We choose the CL framework for finding putative label errors because it was empirically found to outperform several recent alternative label error identification methods 
\cite{northcutt2021confidentlearning, wang2021fair, li2021cleanml}.
Unlike prior work that only models symmetric label noise \citep{rooyen_menon_unhinged_nips15}, we quantify class-conditional label noise with CL, validating the correctable nature of the label errors via crowdsourced workers. Human validation confirms the noise in common benchmark datasets is indeed primarily systematic mislabeling, not just random noise or lack of signal (e.g.\ images with fingers blocking the camera).

\section{Identifying label errors in benchmark datasets}

Here we summarize our algorithmic label error identification performed prior to crowd-sourced human verification. An overview of each dataset and any modifications is detailed in Appendix~\ref{app:datasets}. Step-by-step instructions to obtain each dataset and reproduce the label errors for each dataset are provided at \url{https://github.com/cleanlab/label-errors}. Our code relies on the implementation of confident learning open-sourced at \url{https://github.com/cleanlab/cleanlab}. The primary contribution of this section is not in the methodology, which is covered extensively in \citet{northcutt2021confidentlearning}, but in its utilization as a \emph{filtering} process to significantly (often as much as 90\%) reduce the number of examples requiring  human validation in the next step.

To identify label errors in a test dataset with $n$ examples and $m$ classes, we first characterize label noise in the dataset using the confident learning (CL) framework \citep{northcutt2021confidentlearning} to estimate $\joint$, the $m \times m$ discrete joint distribution of observed, noisy labels, $\tilde{y}$, and unknown, true labels, $y^*$. Inherent in $\joint$ is the assumption that noise is class-conditional \citep{angluin1988learning}, depending only on the latent true class, not the data. This assumption is commonly used \citep{DBLP:conf/iclr/GoldbergerB17_smodel,Sukhbaatar_fergus_iclr_2015, northcutt2017rankpruning} because it is reasonable. For example, in ImageNet, a \emph{tiger} is more likely to be mislabeled \emph{cheetah} than \emph{CD player}.

The diagonal entry $\hat{p}(\tilde{y} \smalleq i , y^* \smalleq i )$ of matrix $\joint$ is the probability that examples in class $i$ are correctly labeled. If the dataset is error-free, then $\sum_{i \in [m]} \hat{p}(\tilde{y} \smalleq i , y^* \smalleq i ) = 1$. The fraction of label errors is $\rho = 1 - \sum_{i \in [m]} \hat{p}(\tilde{y} \smalleq i , y^* \smalleq i )$ and the number of label errors is $\rho \cdot n$. To find label errors, we choose the top $\rho \cdot n$ examples ordered by the normalized margin: $\hat{p}(\tilde{y} \smalleq i; \bm{x}) - \max_{j \neq i} \hat{p}(\tilde{y} \smalleq j; \bm{x})$ \citep{wei2018nomralizedmaxmargin}. Table \ref{tab:datasets} shows the number of CL-guessed label errors for each test set in our study.
CL estimation of $\joint$ is summarized in Appendix~\ref{confidentlearning}.

\begin{table*}[!b]
    \vskip -0.1in
    \caption{Test set errors are prominent across common benchmark datasets. We observe that error rates vary across datasets, from 0.15\% (MNIST) to 10.12\% (QuickDraw); unsurprisingly, simpler datasets, datasets with more carefully designed labeling methodologies, and datasets with more careful human curation generally had less error than datasets that used more automated data collection procedures.}
    \vskip -0.1in
    \label{tab:datasets}
    \center
    \resizebox{\linewidth}{!}{ %Completely zooms in or zooms out (shrinks) entire table!
        \begin{tabular}{l l r l r r r r r}
        \toprule
        \multicolumn{1}{c}{\multirow{2}{*}{\textbf{Dataset}}} &
        \multicolumn{1}{c}{\multirow{2}{*}{\textbf{Modality}}} &
        \multicolumn{1}{c}{\multirow{2}{*}{\textbf{Size}}} &
        \multicolumn{1}{c}{\multirow{2}{*}{\textbf{Model}}} &
        \multicolumn{5}{c}{\textbf{Test Set Errors}} \\
        & & & & \multicolumn{1}{c}{CL guessed} &
        \multicolumn{1}{c}{MTurk checked} &
        \multicolumn{1}{c}{validated} &
        \multicolumn{1}{c}{estimated} &
        \multicolumn{1}{c}{\% error} \\
        \midrule
        MNIST & image & 10,000 & 2-conv CNN & 100 & 100 (100\%) & 15 & - & 0.15\\
        CIFAR-10 & image & 10,000 & VGG & 275 & 275 (100\%) & 54 & - & 0.54 \\
        CIFAR-100 & image & 10,000 & VGG & 2,235 & 2,235 (100\%) & 585 & - & 5.85\\
        Caltech-256$^\dagger$ & image & 29,780 & Wide ResNet-50-2 & 2,360 & 2,360 (100\%) & 458 & - & 1.54 \\
        ImageNet\textsuperscript{*} & image & 50,000 & ResNet-50 & 5,440 & 5,440 (100\%) & 2,916 & - & 5.83 \\
        QuickDraw$^\dagger$ & image & 50,426,266 & VGG & 6,825,383 & 2,500 (0.04\%) & 1870 & 5,105,386 & 10.12 \\
        20news & text & 7,532 & TFIDF + SGD & 93 & 93 (100\%) & 82 & - & 1.09 \\
        IMDB & text & 25,000 & FastText & 1,310 & 1,310 (100\%) & 725 & - & 2.90 \\
        Amazon Reviews$^\dagger$ & text & 9,996,437 & FastText & 533,249 & 1,000 (0.2\%) & 732 & 390,338 & 3.90\\
        AudioSet & audio & 20,371 & VGG & 307 & 307 (100\%) & 275 & - & 1.35\\
        \bottomrule
        \multicolumn{9}{l}{\textsuperscript{*}\footnotesize{Because the ImageNet test set labels are not publicly available, the ILSVRC 2012 validation set is used.}} \\
        \multicolumn{9}{l}{\textsuperscript{$^\dagger$}\footnotesize{Because no explicit test set is provided, we study the entire dataset to ensure coverage of any train/test split.}} \\
        \end{tabular}
    }
\end{table*}

\textbf{Computing out-of-sample predicted probabilities} \; Estimating $\joint$ for CL noise characterization requires two inputs for each dataset: (1) out-of-sample predicted probabilities $\probmatrix$ ($n \times m$ matrix) and (2) the test set labels $\tilde{y}_k$. We observe the best results computing $\probmatrix$ by pre-training on the train set, then fine-tuning (all layers) on the test set using cross-validation to ensure $\probmatrix$ is out-of-sample. If pre-trained models are open-sourced (e.g. ImageNet), we use them instead of pre-training ourselves. If the dataset did not have an explicit test set (e.g. QuickDraw and Amazon Reviews), we skip pre-training and compute $\probmatrix$ using cross-validation on the entire dataset. For all datasets, we try common models that achieve reasonable accuracy with minimal hyper-parameter tuning and use the model yielding the highest cross-validation accuracy, reported in Table \ref{tab:datasets}.

Using this approach allows us to find label errors without manually checking
the entire test set, because CL identifies potential label errors
automatically.

\section{Validating label errors with Mechanical Turk}
\label{validating}

We validated the algorithmically identified label errors with a Mechanical Turk (MTurk) study.
For two large datasets with a large number of errors (QuickDraw and Amazon Reviews), we checked a random sample; for the rest, we checked all identified errors.

We presented workers with hypothesized errors and asked them whether they saw the (1) given label, (2) the top CL-predicted label, (3) both labels, or (4) neither label in the example. To aid the worker, the interface included high-confidence examples of the given class and the CL-predicted class. Figure~\ref{fig:mturk-interface} in Appendix~\ref{sec:mturk-interface} shows a screenshot of the MTurk worker interface.

Each CL-flagged label error was independently presented to five workers. We consider the example validated (an ``error'') if fewer than three of the workers agree that the data point has the given label (\emph{agreement threshold = 3 of 5}) , otherwise we consider it to be a ``non-error'' (i.e.\ the original label was correct).
We further categorize the label errors, breaking them down into (1) ``correctable'', where a majority agree on the CL-predicted label; (2) ``multi-label'', where a majority agree on both labels appearing; (3) ``neither'', where a majority agree on neither label appearing; and (4) ``non-agreement'', a catch-all category for when there is no majority.
Table~\ref{tab:mturk} summarizes the results, and Figure~\ref{fig:image-val} shows examples of validated label errors from image datasets.

\begin{table}
    \vskip -0.3in
\setlength\tabcolsep{2pt} % Makes table columns tighter
    \caption{Mechanical Turk validation of CL-flagged errors and categorization of label issues.}
    \label{tab:mturk}
    \center
    
\resizebox{.65\linewidth}{!}{ %Completely zooms in or zooms out (shrinks) entire table!
        \begin{tabular}{l r r r r r r}
        \toprule
        \multicolumn{1}{c}{\multirow{2}{*}{\textbf{Dataset}}} &
        \multicolumn{6}{c}{\textbf{Test Set Errors Categorization}} \\
        & \multicolumn{1}{c}{non-errors} &
        \multicolumn{1}{c}{errors} &
        \multicolumn{1}{c}{non-agreement} &
        \multicolumn{1}{c}{correctable} &
        \multicolumn{1}{c}{multi-label} &
        \multicolumn{1}{c}{neither} \\
        \midrule
        MNIST & 85 & 15 & 2 & 10 & - & 3 \\
        CIFAR-10 & 221 & 54 & 32 & 18 & 0 & 4 \\
        CIFAR-100 & 1650 & 585 & 210 & 318 & 20 & 37 \\
        Caltech-256 & 1902 & 458 & 99 & 221 & 115 & 23 \\
        ImageNet & 2524 & 2916 & 598 & 1428 & 597 & 293 \\
        QuickDraw & 630 & 1870 & 563 & 1047 & 20 & 240 \\
        20news & 11 & 82 & 43 & 22 & 12 & 5 \\
        IMDB & 585 & 725 & 552 & 173 & - & - \\
        Amazon Reviews & 268 & 732 & 430 & 302 & - & - \\
        AudioSet & 32 & 275 & - & - & - & - \\
        \bottomrule
        \end{tabular}
}
\end{table}

\subsection{Failure modes of confident learning}

\begin{figure*}[!h]
\centerline{\includegraphics[width=\textwidth]{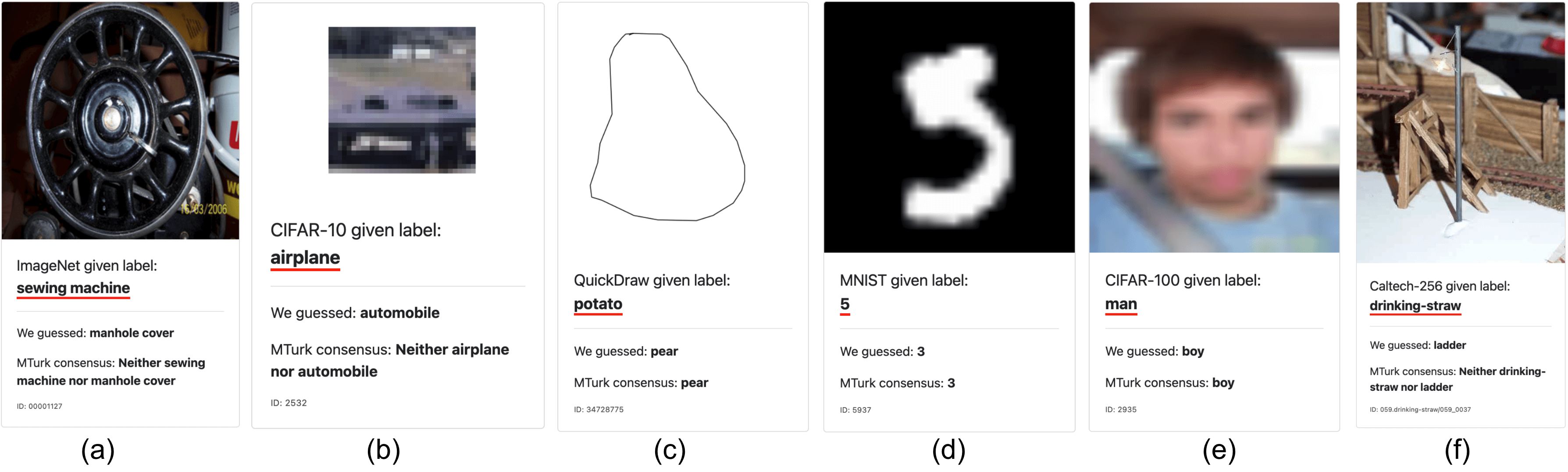}}
\vskip -0.05in
\caption{Difficult examples from various datasets where confident learning finds a potential label error but human validation shows that there actually is no error. Example (a) is a cropped image of part of an antiquated sewing machine; (b) is a viewpoint from inside an airplane, looking out at the runway and grass with a partial view of the nose of the plane; (c) is an ambiguous shape which could be a potato; (d) is likely a badly drawn ``5''; (e) is a male whose exact age cannot be determined; and (f) is a straw used as a pole within a miniature replica of a village.}
\label{fig:failure_mode}
\vskip -.05in
\end{figure*}

Confident learning sometimes flags data points that are not actually erroneous.
By visually inspecting putative label errors, we identified certain previously unexamined failure modes of confident learning  \citep{northcutt2021confidentlearning}. Appendix \ref{sec:failure_modes_math} provides a mathematical description of the conditions under which these failure modes occur. 
Figure \ref{fig:failure_mode} shows uniquely challenging examples, with excessively erroneous $\hat{p}(\tilde{y} \smalleq j; \bm{x})$, where failure mode cases potentially occur. The sewing machine in Figure \ref{fig:failure_mode}(a), for example, exhibits a ``part versus whole'' issue where the image has been cropped to only show a small portion of the object. The airplane in Figure \ref{fig:failure_mode}(b) is an unusual example of the class, showing the plane from the perspective of the pilot looking out of the front cockpit window.

Figure \ref{fig:failure_mode} clarifies that our corrected test set labels are not 100\% perfect. Even with a stringent 5 of 5 agreement threshold where all human reviewers agreed on a label correction, the ``corrected'' label is not always actually correct. Fortunately, these failure mode cases are rare. Inspection of \url{https://labelerrors.com} shows that the majority of the labels we corrected are reasonable. Our corrected test sets, while imperfect in these cases, are improved from the original test sets.

\section{Implications of label errors in test data} \label{sec:implications_discussion}

Finally, we consider how pervasive test set label errors may affect ML practitioners in real-world applications. 
To clarify the discussion, we first introduce some useful terminology.

\begin{definition}[original accuracy, $\tilde{A}$]
  The accuracy of a model's predicted labels over a given dataset, computed with respect to the original labels present in the dataset. Measuring $\tilde{A}$ over the test set is the standard way practitioners evaluate their models today.
\end{definition}

\begin{definition}[corrected accuracy, $A^*$]
  The accuracy of a model's predicted labels, computed over a modified dataset in which previously identified erroneous labels have been corrected by humans to the true class when possible and removed when not.
  Measuring $A^*$ over the test set is preferable to $\tilde{A}$ for evaluating models because $A^*$ better reflects performance in real-world applications.
\end{definition}

The \emph{human} labelers referenced throughout this section are the workers from our MTurk study in Section  \ref{validating}. 
In the following definitions, $\backslash$ denotes a set difference and $\mathcal{D}$ denotes the full test dataset.

\begin{definition}[benign set, $\mathcal{B}$]
    The subset of benign test examples, comprising data that CL did not flag as likely label errors and data that was flagged but for which human reviewers agreed that the original label should be kept. ($\mathcal{B} \subset \mathcal{D}$)
\end{definition}
 
\begin{definition}[unknown-label set, $\mathcal{U}$]
    The subset of CL-flagged test examples for which human labelers could not agree on a single correct label. This includes examples where human reviewers agreed that multiple classes or none of the classes are appropriate. ($\mathcal{U} \subset \mathcal{D} \backslash \mathcal{B}$)
\end{definition}

\begin{definition}[pruned set, $\mathcal{P}$]
    The remaining test data after removing $\mathcal{U}$ from  $\mathcal{D}$. ($\mathcal{P} = \mathcal{D} \backslash \mathcal{U}$)
\end{definition}

\begin{definition}[correctable set, $\mathcal{C}$]
    The subset of CL-flagged examples for which human-validation reached consensus on a different label than the originally given label. ($\mathcal{C} = \mathcal{P} \backslash \mathcal{B}$)
\end{definition} 

\begin{definition}[noise prevalence, $N$]
    The percentage of the pruned set comprised of the correctable set, i.e.\ what fraction of data received the wrong label in the original benchmark when a clear alternative ground-truth label should have been assigned  (disregarding any data for which humans failed to find a clear alternative). Here we operationalize noise prevalence as $N = \frac{\vert \mathcal{C} \vert}{\vert \mathcal{P} \vert}$.
\end{definition}

These definitions imply $\mathcal{B}, \mathcal{C}, \mathcal{U}$ are disjoint with  $\mathcal{D} = \mathcal{B} \cup \mathcal{C} \cup \mathcal{U}$ and also   $\mathcal{P} = \mathcal{B} \cup \mathcal{C}$. In subsequent experiments, we report corrected test accuracy over $\mathcal{P}$ after correcting all of the labels in $\mathcal{C} \subset \mathcal{P}$. 
We ignore the unknown-label set $\mathcal{U}$ (and do not include those examples in our estimate of noise prevalence)  because it is unclear how to measure \emph{corrected accuracy} for examples whose true underlying label remains ambiguous. Thus the \emph{noise prevalence} reported throughout this section differs from the fraction of label errors originally found in each of the test sets.

A major issue in ML today is that one only sees the original test accuracy in practice, whereas one would prefer to base modeling decisions on the corrected  test accuracy instead. 
Our subsequent discussion highlights the potential implications of this mismatch. 
What are the consequences of test set label errors? Figure~\ref{fig:imagenet_benchmarking} compares performance on the ImageNet validation set, \textit{commonly used in place of the test set}, of 34 pre-trained models from the PyTorch and Keras repositories (throughout, we use provided checkpoints of models that have been fit to the original training set). Figure~\ref{fig:orig_vs_clean} confirms the observations of \citet{recht2019imagenet}; benchmark conclusions are largely unchanged by using a corrected test set, i.e.\ in our case by removing errors.

\subsection{Benchmarking on the correctable set}
\label{sec:correctable}

\begin{figure*}[t]
% \vskip -0.1in
\begin{subfigure}{0.33\textwidth}
\includegraphics[width=0.98\linewidth,]{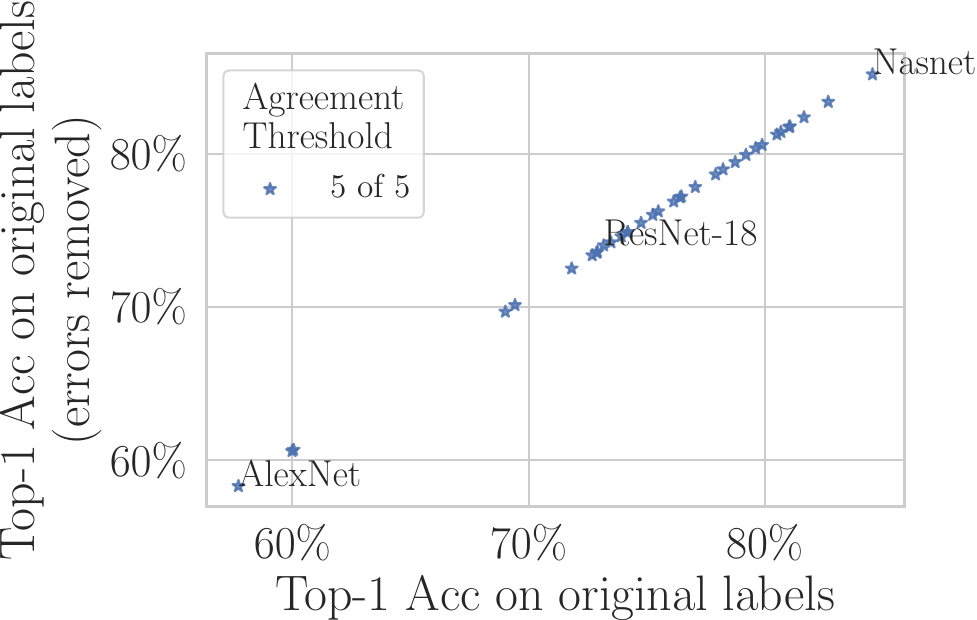} 
\vskip -0.05in
\caption{ImageNet val set acc.}
\label{fig:orig_vs_clean}
\end{subfigure}
\begin{subfigure}{0.33\textwidth}
\includegraphics[width=0.98\linewidth, ]{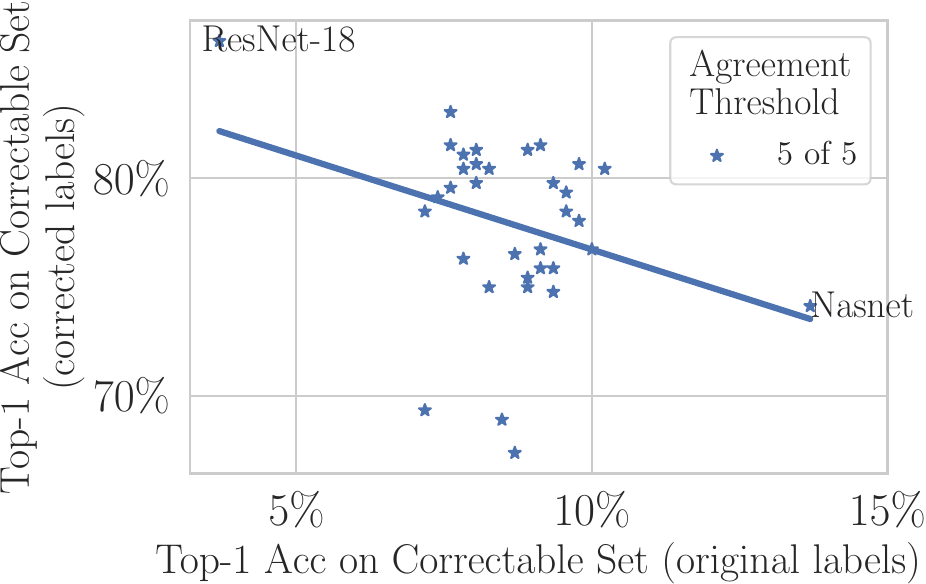}
\vskip -0.05in
\caption{ImageNet correctable set acc.}
\label{fig:orig_vs_corrected}
\end{subfigure}
\begin{subfigure}{0.33\textwidth}
\includegraphics[width=0.98\linewidth, ]{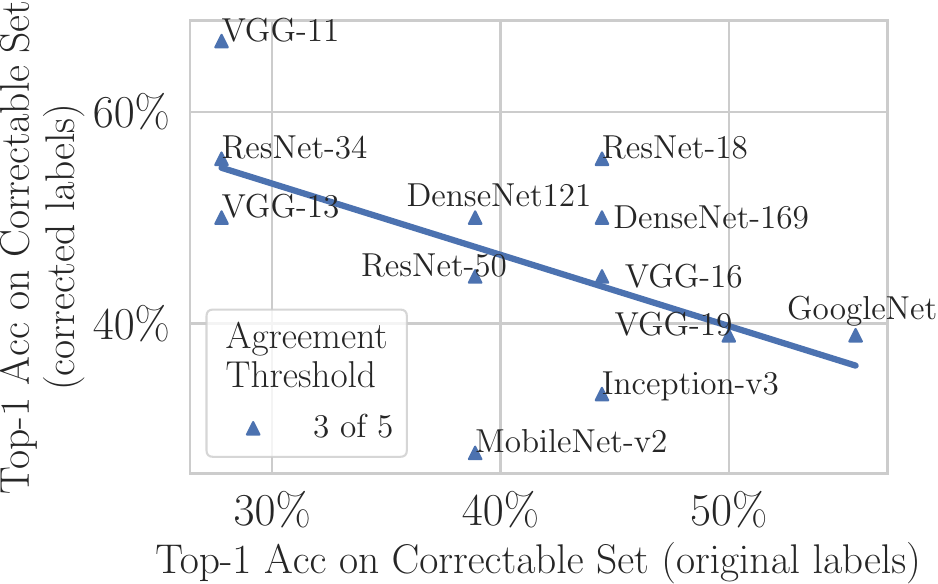}
\vskip -0.05in
\caption{CIFAR-10 correctable set acc.}
\label{fig:cifar10_orig_vs_corrected}
\end{subfigure} 
    \caption{Benchmark ranking comparison of 34 pre-trained ImageNet models and 13 pre-trained CIFAR-10 models (details in Tables \ref{tab:cifar10_benchmarking_table} and \ref{tab:imagenet_benchmarking_table} and Figure \ref{fig:orig_vs_corrected_all} in the Appendix). Benchmarks are unchanged by removing label errors (a), but change drastically (b) on the Correctable set with original (erroneous) labels versus corrected labels, e.g.\ NASNet: 1/34 $\rightarrow$ 29/34, ResNet-18: 34/34 $\rightarrow$ 1/34. 
}
\label{fig:imagenet_benchmarking}
\vskip -0.1in
\end{figure*}

However, we find a surprising result upon closer examination of the models' performance on the correctable set $\mathcal{C}$.
When evaluating models \emph{only} on these originally-mislabeled test data, models which perform best on the original (incorrect) labels perform the worst on the corrected labels. For example, ResNet-18 \citep{he2016deep} significantly outperforms NASNet \citep{zoph2018learning} in terms of corrected accuracy over $\mathcal{C}$, despite exhibiting far worse original test accuracy. The change in ranking can be dramatic: NASNet-large drops from ranking 1/34 $\rightarrow$ 29/34, Xception drops from ranking 2/34 $\rightarrow$ 24/34, ResNet-18 increases from ranking 34/34 $\rightarrow$ 1/34, and ResNet-50 increases from ranking 20/24 $\rightarrow$ 2/24 (see Table \ref{tab:imagenet_benchmarking_table} in the Appendix).  
We verified that the same trend occurs independently across 13 pre-trained CIFAR-10 models (Figure~\ref{fig:cifar10_orig_vs_corrected}), e.g. VGG-11 significantly outperforms VGG-19 \citep{simonyan2014vgg} in terms of corrected accuracy over $\mathcal{C}$. Note that all numbers reported here are  over subsets of the held-out test data, so this is not overfitting in the classical sense.

This phenomenon, depicted in Figures~\ref{fig:orig_vs_corrected} and \ref{fig:cifar10_orig_vs_corrected}, may indicate two key insights: (1) lower-capacity models provide unexpected regularization benefits and are more resistant to learning the asymmetric distribution of noisy labels, 
(2) over time, the more recent (larger) models have architecture/hyperparameter decisions that were made on the basis of the (original) test accuracy. Learning to capture systematic patterns of label error in their predictions allows these models to improve their original test accuracy, but this is not the sort of progress ML research should aim to achieve. 
\citet{harutyunyan2020improving} and \citet{arpit2017closer} have previously analyzed phenomena similar to (1), and here we demonstrate that this issue really does occur for the models/datasets widely used in current practice.
(2) is an undesirable form of overfitting, albeit not in the classical sense (as the original test accuracy can further improve through better modeling of label errors), but rather overfitting to the specific benchmark (and quirks of the original label annotators) such that accuracy improvements for erroneous labels may not translate to superior performance in a deployed ML system.

This phenomenon has important practical implications for real-world datasets with greater noise prevalence than the highly curated benchmark data studied here. 
In these relatively clean benchmark datasets, the noise prevalence is an underestimate as we could only verify a subset of our candidate label errors rather than all test labels, and thus the potential gap between original vs.\ corrected test accuracy over $\mathcal{P}$ is limited for these particular benchmarks. 
However, this gap increases proportionally for data with more (correctable) label errors in the test set, i.e.\ as $N$ increases.

\begin{figure*}[t]
\vskip 0in
\centering
\hspace*{.2in}\includegraphics[width=0.88\linewidth, ]{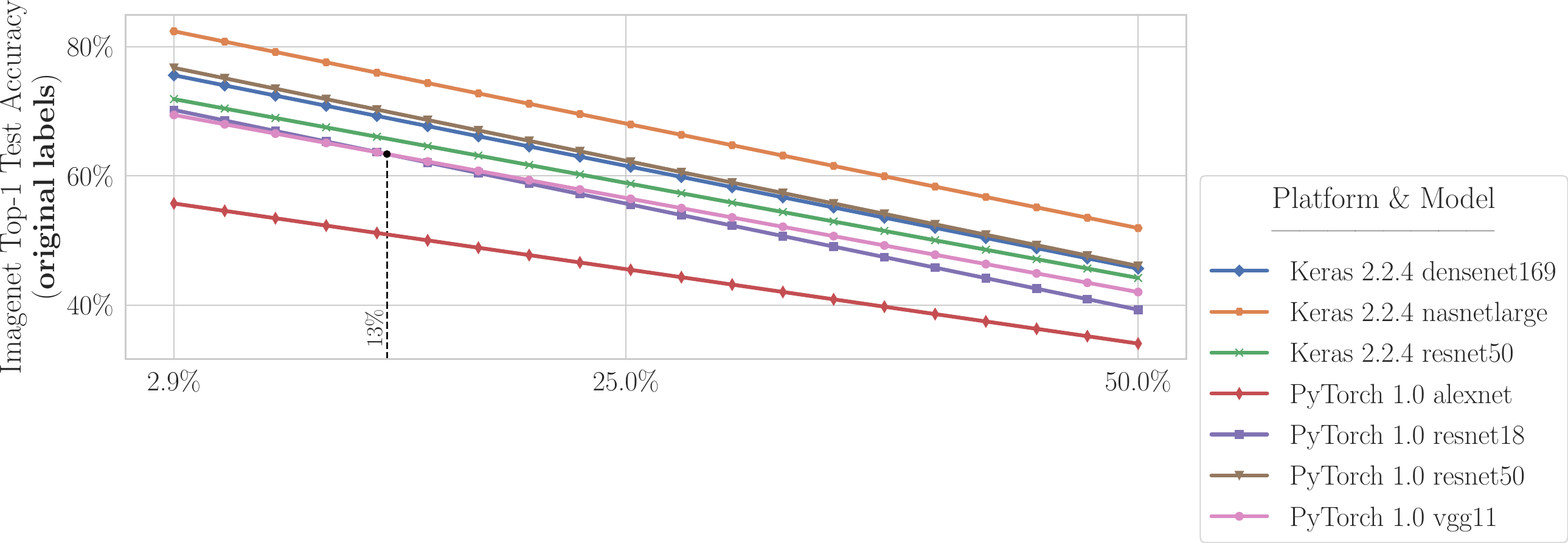}
\vskip -0.47in
\hspace*{.2in}\hspace*{-1.188in}\includegraphics[width=0.666\linewidth, ]{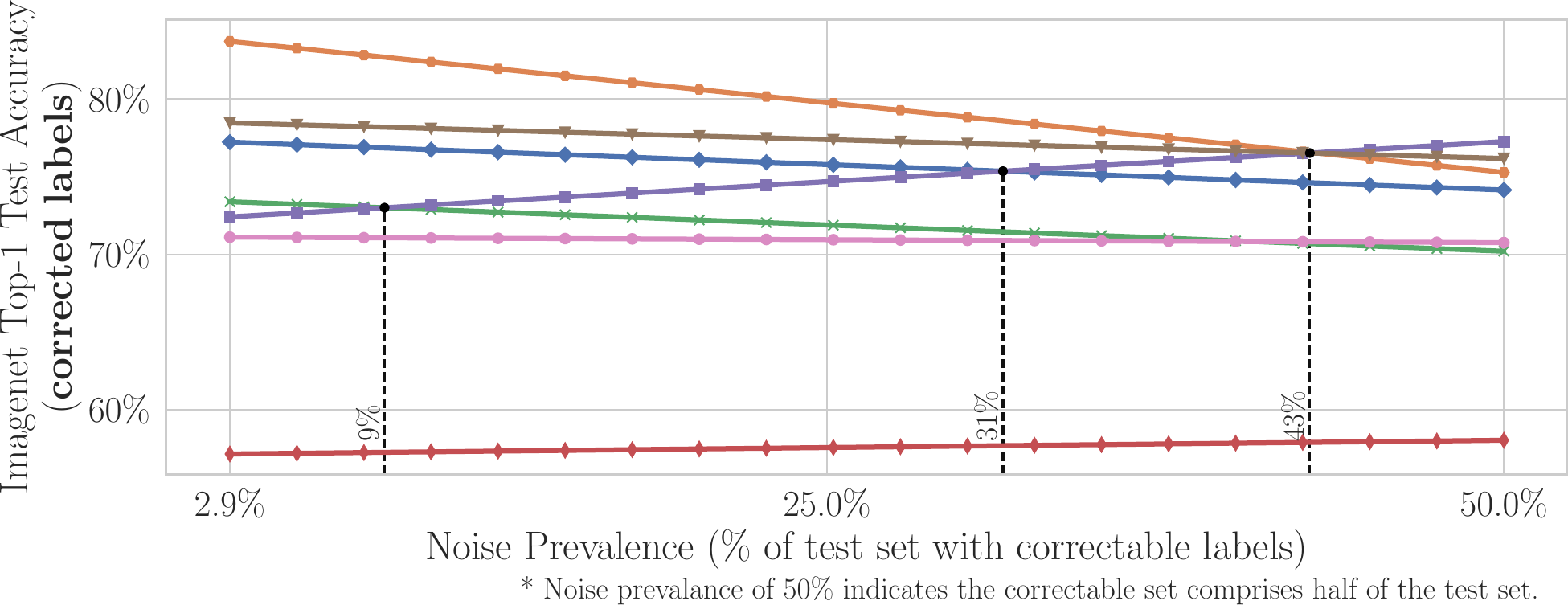}
    \caption{ImageNet top-1 original accuracy (top) and corrected accuracy (bottom) vs noise prevalence (agreement threshold = 3). Vertical lines indicate noise levels at which the ranking of two models changes (in terms of original/corrected accuracy). The left-most point ($N = 2.9$\%) on the x-axis is $|\mathcal{C}|/|\mathcal{P}|$, i.e. the (rounded) estimated noise prevalence of the pruned set, $\mathcal{P}$. The leftmost vertical dotted line in the bottom panel is read, ``The ResNet-50 and ResNet-18 benchmarks cross at noise prevalence $N = 9\%$,'' implying ResNet-18 outperforms ResNet-50 when $N$ increases by around $6\%$ relative to the original pruned test data ($N = 2.9\%$ originally, c.f.\ Table \ref{tab:mturk}).
}
\label{fig:imagenet_original_noise_prevalance_agreement_threshold_3}
\vskip 0.1in
\end{figure*}

\begin{figure*}[!t]
\vskip -0.05in
\centering
\hspace*{.2in}\includegraphics[width=0.88\linewidth, ]{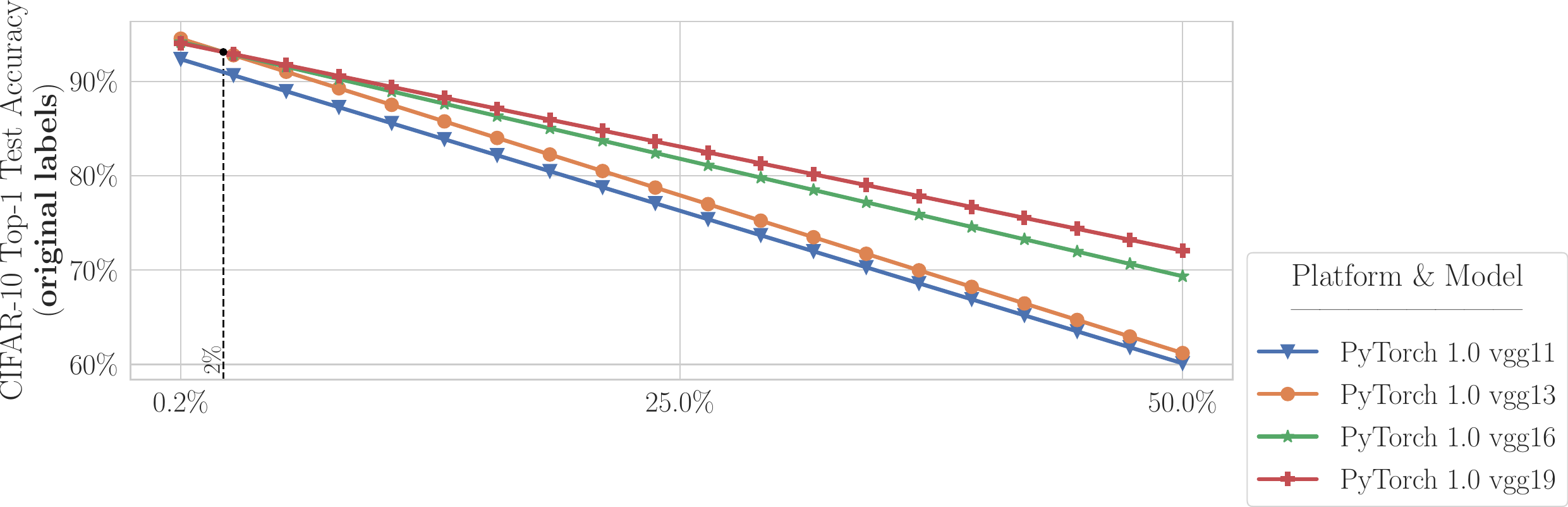}
\vskip -0.3in
\hspace*{.2in}\hspace*{-1.03in}\includegraphics[width=0.6923\linewidth, ]{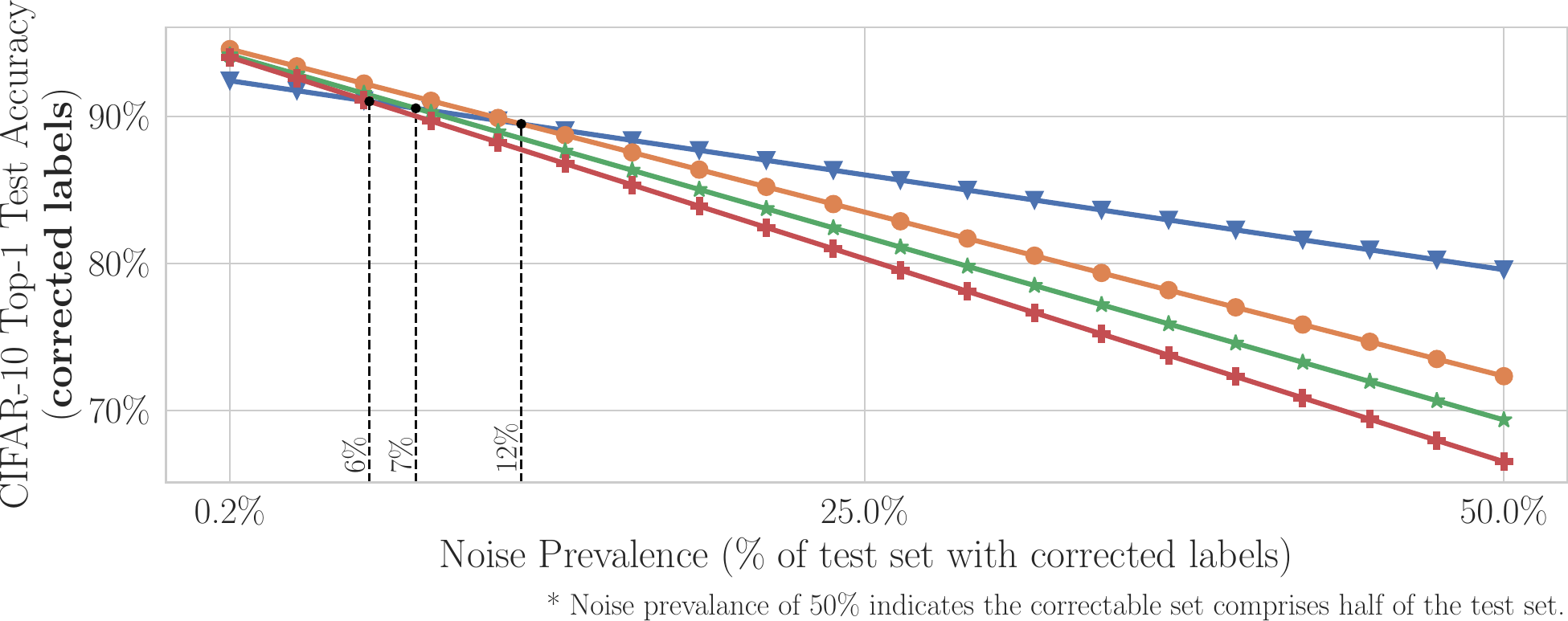}
\caption{CIFAR-10 top-1 original accuracy (top panel) and corrected accuracy (bottom panel) vs Noise Prevalence (agreement threshold = 3). For additional details, see the caption of Figure \ref{fig:imagenet_original_noise_prevalance_agreement_threshold_3}.
}
\label{fig:cifar10_original_noise_prevalance_agreement_threshold_3}
\end{figure*}

\subsection{Benchmark instability}

To investigate how benchmarks of popular models change with varying proportions  of label errors in test sets, we randomly and incrementally remove correctly-labeled examples, one at a time, until only the original set of mislabeled test data (with corrected labels) is left. We create alternate versions (subsets) of the pruned benchmark test data $\mathcal{P}$, in which we additionally randomly omit some fraction, $x$, of $\mathcal{B}$ (the non-CL-flagged test examples). This effectively increases the proportion of the resulting test dataset comprised of the correctable set $\mathcal{C}$, and reflects how test sets function in applications with greater prevalence of label errors. If we remove a fraction $x$ of benign test examples (in $\mathcal{B}$) from $\mathcal{P}$, we estimate the noise prevalence in the new (reduced) test dataset to be $N = \frac{|\mathcal{C}|}{|\mathcal{P}| - x|\mathcal{B}|}$. By varying $x$ from $0$ to $1$, we can simulate any noise prevalence ranging from $|\mathcal{C}|/|\mathcal{P}|$ to $1$. We operationalize averaging over all choices of removal by linearly interpolating from accuracies over the corrected test set ($\mathcal{P}$, with corrected labels for the subset $\mathcal{C}$) to accuracies over the erroneously labeled subset ($\mathcal{C}$, with corrected labels). Over these corrected test sets, we evaluate popular pre-trained models (again using provided checkpoints of models that have been fit to the original training set).

For a given model $\mathcal{M}$, its resulting accuracy (as a function of $x$) over the reduced test data is given by $A(x; \mathcal{M}) = \frac{A_\mathcal{C}(\mathcal{M}) \cdot |\mathcal{C}| + (1-x) \cdot A_\mathcal{B}(\mathcal{M}) \cdot |B|}{|\mathcal{C}| + (1-x) \cdot |B|}$, where $A_\mathcal{C}(\mathcal{M})$ and $A_\mathcal{B}(\mathcal{M})$ denote the (original or corrected) accuracy over the correctable set and benign set, respectively (accuracy before removing any examples).
Here $A_\mathcal{B} = A_\mathcal{B}^* = \tilde{A}_\mathcal{B}$ because no erroneous 
labels were identified in $\mathcal{B}$.
The expectation is taken over which fraction $x$ of examples are randomly removed from $\mathcal{B}$ to produce the reduced test set: the resulting expected accuracy, $A(x; \mathcal{M})$, is depicted on the y-axis of Figures \ref{fig:imagenet_original_noise_prevalance_agreement_threshold_3}-\ref{fig:cifar10_original_noise_prevalance_agreement_threshold_3}. As our removal of test examples was random from the non-mislabeled set, we expect this reduced test data is representative of test sets that would be used in applications with a similarly greater prevalence of label errors.
Note that we ignore non-correctable data with unknown labels ($\mathcal{U}$) throughout this analysis, as it is unclear how to report a better version of the accuracy for such ill-specified examples.

Over alternative (reduced) test sets created by imposing increasing degrees of noise prevalence in ImageNet/CIFAR-10, 
Figures \ref{fig:imagenet_original_noise_prevalance_agreement_threshold_3}-\ref{fig:cifar10_original_noise_prevalance_agreement_threshold_3} depict the resulting original (erroneous) test set accuracy and corrected accuracy of the models, expected on each alternative test set. For a given test set (i.e.\ point along the $x$-axis of these plots), the vertical ordering of the lines indicates how models would be favored based on original accuracy or corrected accuracy over this test set. Unsurprisingly, we see that more flexible/recent architectures tend to be favored on the basis of original accuracy, regardless of which test set (of varying noise prevalence) is considered. 
This aligns with conventional expectations that powerful models like NASNet will outperform simpler models like ResNet-18. 
However, if we shift our focus to the corrected accuracy (i.e.\ what actually matters in practice), it is no longer the case that more powerful models are reliably better than their simpler counterparts: the performance strongly depends on the degree of noise prevalence in the test data. For datasets where label errors are common, a practitioner is more likely to select a model (based on original accuracy) that is not actually the best model (in terms of corrected accuracy) to deploy.

Finally, we note that this analysis only presents a loose lower bound on the magnitude of these issues due to unaccounted for label errors in the non-CL-flagged data (see Section \ref{sec:expertreview}).
We only identified a subset of the actual correctable set as we are limited to human-verifiable label corrections for a subset of data candidates (algorithmically prioritized via confident learning). Because the actual correctable sets are likely larger, our noise prevalence estimates are optimistic in favor of higher capacity models. Thus, 
the true gap between corrected vs.\ original accuracy may be larger and of greater practical significance, even for the gold-standard benchmark datasets considered here. For many application-specific datasets collected by ML practitioners, the noise prevalence will be greater than the numbers presented here: thus, it is imperative to be cognizant of the distinction between corrected vs.\ original accuracy, and to utilize careful data curation practices, perhaps by allocating more of an annotation budget to ensure higher quality labels in the test data.

\section{Expert review of CL-flagged and non-CL-flagged label errors in ImageNet}
\label{sec:expertreview}

Up to this point, we have only evaluated the subsets of the datasets flagged by CL: how do we know that CL-flagged examples are indeed more erroneous than a random subset of a dataset? How many label errors are missed in the non-CL-flagged data? And how reliable are MTurk workers in comparison to expert reviewers? In this section, we address these questions by conducting an additional expert review of both CL-flagged and non-CL-flagged examples in the ImageNet val set.

The expert review was conducted in two phases (details in Appendix \ref{sec:expertdetails}). In the first phase, experts reviewed 1 randomly-selected CL-flagged example and 1 randomly-selected non-CL-flagged example from each of the 1,000 ImageNet classes (66 classes had no CL-flagged example, i.e. 934 + 1,000 = 1934 images were evaluated in total). Given a similar interface as MTurk workers, the expert reviewers selected one choice from: (1) the given label, (2) the top-most predicted label that differs from the given label, (3) ``both'', and (4) ``neither''.
Experts researched any unfamiliar classes by looking up related images and taxonomy information online, spending on average 13x more time per label than MTurk workers. 
Each image was reviewed by at least two experts, and experts agreed on decisions for 77\% of the images. In the second phase, all experts jointly revisited the remaining 23\% where there was disagreement and came to a consensus on a single choice.

Table \ref{tab:expert_percent} reveals that the set of CL-flagged examples has significantly higher proportions of every type of label issue than the set of non-CL-flagged examples. 
An image flagged by CL was 2.6x as likely to be erroneously labeled than an non-CL-flagged image. 
Given a limited budget for human review, we thus recommend using CL to prioritize examples when verifying the labels in a large dataset. 
 
\begin{table}[t!]
    \caption{Percentages of label errors identified by experts vs.\ MTurk workers in CL-flagged examples and random non-CL-flagged examples from ImageNet. Only experts reviewed non-CL examples. The first two rows are computed over the same subset of images. The last column lists average time spent reviewing each image.  Percentages are row-normalized, with raw counts provided in Table \ref{tab:expert_count}.} 
\label{tab:expert_percent}
\center
\begin{tabular}{lrrrrr|r}
\toprule
{} & non-errors & errors & correctable & multi-label & neither & Avg. time spent \\
\midrule
CL (MTurk)      &      57.9\% &  42.2\% &       24.7\% &       11.1\% &    6.4\% & 5 seconds \\
CL (expert)     &      58.7\% &  41.4\% &       17.7\% &       13.1\% &   10.6\% & 67 seconds\\
non-CL (expert) &      84.0\% &  16.0\% &        3.2\% &        9.1\% &    3.7\% & 67 seconds\\
\bottomrule
\end{tabular}
\vskip -.1in
\end{table}

Comparing \emph{CL (expert)} to \emph{CL (MTurk)} in Table \ref{tab:expert_percent} indicates that for CL-flagged examples, MTurk workers favored correcting labels in cases where experts agreed neither label was appropriate. For this analysis, we only consider the subset of MTurk reviewed images that overlaps with the 1,934 expert reviewed images. This may be attributed to experts knowing a better choice than the two label choices presented in the task (c.f.\ Figure~\ref{fig:mturk-mistake}).
Nonetheless the MTurk results overall agree with those from our expert review. This validates our overall approach of using CL followed by MTurk to characterize label errors, and demonstrates that a well-designed interface (Figure \ref{fig:mturk-interface}) suffices for non-expert workers to provide high-quality label verification of datasets.

We further estimate that the analysis in previous sections missed around 14\% of the label errors in ImageNet because 89\% of images were not flagged by CL and Table \ref{tab:expert_percent} indicates around 16\% of these were mislabeled. By including the additional 14\% error found from the \emph{9x larger} set of non-CL-flagged examples, we can more accurately estimate that the ImageNet validation set contains closer to 20\% label errors (up from the 6\% reported in Table \ref{tab:datasets}). This roughly indicates \emph{how much more} severe the issue of label errors actually is compared to what we reported in Sections \ref{validating} and \ref{sec:implications_discussion}.

\section{Discussion} \label{sec:discussion}

This paper demonstrates that label errors are ubiquitous in the test sets of many popular benchmarks used to gauge progress in machine learning. 
We hypothesize that this has not been previously discovered and publicized at such scale due to various challenges. Firstly, human verification of all labels can be quite costly, which we circumvented here by using CL algorithms to filter automatically for likely label errors prior to human verification. Secondly, working with 10 differently formatted  datasets was nontrivial, with some exhibiting peculiar issues upon close inspection (despite being standard benchmarks). For example, IMDB, QuickDraw, and Caltech-256 lack a global index making it difficult to map model outputs to corrected test examples on different systems. We provide index files in our repository\footnote{\url{https://github.com/cleanlab/label-errors\#how-to-download-prepare-and-index-the-datasets}} to address this.  
Furthermore, Caltech-256 contains several duplicate images, of which which we found no previous mention. 
Lastly, ImageNet contains duplicate class labels, e.g.\ ``maillot'' (638 \& 639) and ``crane''  (134 \& 517) \citep{tsipras2020imagenet, northcutt2021confidentlearning}.

Traditionally, ML practitioners choose which model to deploy based on
test accuracy — our findings advise caution here. 
Instead, judging models over correctly labeled
test sets may be important, especially for 
 real-world datasets that are likely noisier than these popular benchmarks. Small increases in the prevalence of mislabeled test data can destabilize ML benchmarks, indicating that low-capacity models may actually outperform high-capacity models in noisy real-world applications, even if their measured performance on the original test data appears worse. 
We recommend considering the distinction between corrected vs.\ original test accuracy and curating datasets to maximize high-quality test labels, even if budget constraints only allow for lower-quality training labels.

This paper shares new findings about pervasive label errors in test sets and their effects on benchmark stability, but it does not address whether the apparent overfitting of high-capacity models versus low-capacity models is due to overfitting to train set noise, overfitting to validation set noise during hyper-parameter tuning, or heightened sensitivity to train/test label distribution shift that occurs when test labels are corrected.
An intuitive hypothesis is that high-capacity models more closely fit all statistical patterns present in the data, including those patterns related to systematic label errors that models with more limited capacity are less capable of closely approximating. 
A rigorous analysis to disambiguate and understand the contribution of each of these causes and their effects on benchmarking stability is a natural next step, which we leave for future work. How to best allocate a given human label verification budget between training and test data also remains an open question. 

 \section*{Acknowledgments}
% \paragraph{Acknowledgments}
This work was supported in part by funding from the MIT-IBM Watson AI Lab. We thank Jessy Lin for her contributions to early stages of this research, and we thank Wei Jing Lok for his contributions to the ImageNet expert labeling experiments.

\clearpage
\bibliography{paper}

\newcommand{\noopsort}[1]{} \newcommand{\printfirst}[2]{#1}
  \newcommand{\singleletter}[1]{#1} \newcommand{\switchargs}[2]{#2#1}
\begin{thebibliography}{50}
\providecommand{\natexlab}[1]{#1}
\providecommand{\url}[1]{\texttt{#1}}
\expandafter\ifx\csname urlstyle\endcsname\relax
  \providecommand{\doi}[1]{doi: #1}\else
  \providecommand{\doi}{doi: \begingroup \urlstyle{rm}\Url}\fi

\bibitem[Angluin and Laird(1988)]{angluin1988learning}
D.~Angluin and P.~Laird.
\newblock Learning from noisy examples.
\newblock \emph{Machine Learning}, 2\penalty0 (4):\penalty0 343--370, 1988.

\bibitem[Arpit et~al.(2017)Arpit, Jastrz{\k{e}}bski, Ballas, Krueger, Bengio,
  Kanwal, Maharaj, Fischer, Courville, Bengio, et~al.]{arpit2017closer}
D.~Arpit, S.~Jastrz{\k{e}}bski, N.~Ballas, D.~Krueger, E.~Bengio, M.~S. Kanwal,
  T.~Maharaj, A.~Fischer, A.~Courville, Y.~Bengio, et~al.
\newblock A closer look at memorization in deep networks.
\newblock In \emph{International Conference on Machine Learning}, pages
  233--242. Proceedings of Machine Learning Research (PMLR), 2017.

\bibitem[Chen et~al.(2019)Chen, Liao, Chen, and Zhang]{chen2019confusion}
P.~Chen, B.~B. Liao, G.~Chen, and S.~Zhang.
\newblock Understanding and utilizing deep neural networks trained with noisy
  labels.
\newblock In \emph{International Conference on Machine Learning (ICML)}, 2019.

\bibitem[{Cordeiro} and {Carneiro}(2020)]{survery_deep_learning_noisy_labels}
F.~R. {Cordeiro} and G.~{Carneiro}.
\newblock A survey on deep learning with noisy labels: How to train your model
  when you cannot trust on the annotations?
\newblock In \emph{Conference on Graphics, Patterns and Images (SIBGRAPI)},
  pages 9--16, 2020.

\bibitem[Dawid and Skene(1979)]{dawid1979maximum}
A.~P. Dawid and A.~M. Skene.
\newblock Maximum likelihood estimation of observer error-rates using the em
  algorithm.
\newblock \emph{Journal of the Royal Statistical Society: Series C (Applied
  Statistics)}, 28\penalty0 (1):\penalty0 20--28, 1979.

\bibitem[Deng et~al.(2009)Deng, Dong, Socher, Li, Li, and Fei-Fei]{imagenet}
J.~Deng, W.~Dong, R.~Socher, L.-J. Li, K.~Li, and L.~Fei-Fei.
\newblock {ImageNet: A Large-Scale Hierarchical Image Database}.
\newblock In \emph{Conference on Computer Vision and Pattern Recognition
  (CVPR)}, 2009.

\bibitem[Fr{\'{e}}nay and Verleysen(2014)]{Frenay2014}
B.~Fr{\'{e}}nay and M.~Verleysen.
\newblock {Classification in the presence of label noise: A survey}.
\newblock \emph{IEEE Transactions on Neural Networks and Learning Systems},
  25\penalty0 (5):\penalty0 845--869, 2014.
\newblock ISSN 21622388.
\newblock \doi{10.1109/TNNLS.2013.2292894}.

\bibitem[Gemmeke et~al.(2017)Gemmeke, Ellis, Freedman, Jansen, Lawrence, Moore,
  Plakal, and Ritter]{audioset}
J.~F. Gemmeke, D.~P.~W. Ellis, D.~Freedman, A.~Jansen, W.~Lawrence, R.~C.
  Moore, M.~Plakal, and M.~Ritter.
\newblock Audio set: An ontology and human-labeled dataset for audio events.
\newblock In \emph{IEEE International Conference on Acoustics, Speech, and
  Signal Processing (ICASSP)}, New Orleans, LA, 2017.

\bibitem[Goldberger and
  Ben{-}Reuven(2017)]{DBLP:conf/iclr/GoldbergerB17_smodel}
J.~Goldberger and E.~Ben{-}Reuven.
\newblock Training deep neural-networks using a noise adaptation layer.
\newblock In \emph{International Conference on Learning Representations
  (ICLR)}, 2017.

\bibitem[Griffin et~al.(2007)Griffin, Holub, and Perona]{caltech}
G.~Griffin, A.~Holub, and P.~Perona.
\newblock Caltech-256 object category dataset.
\newblock Technical Report 7694, California Institute of Technology, 2007.
\newblock URL \url{http://authors.library.caltech.edu/7694}.

\bibitem[Grother(1995)]{grother1995nist_mnistcuration}
P.~J. Grother.
\newblock Nist special database 19 handprinted forms and characters database.
\newblock \emph{National Institute of Standards and Technology}, 1995.

\bibitem[Ha and Eck(2017)]{quickdraw2017google}
D.~Ha and D.~Eck.
\newblock A neural representation of sketch drawings.
\newblock \emph{arXiv preprint arXiv:1704.03477}, 2017.

\bibitem[Harutyunyan et~al.(2020)Harutyunyan, Reing, Ver~Steeg, and
  Galstyan]{harutyunyan2020improving}
H.~Harutyunyan, K.~Reing, G.~Ver~Steeg, and A.~Galstyan.
\newblock Improving generalization by controlling label-noise information in
  neural network weights.
\newblock In \emph{International Conference on Machine Learning (ICML)}, pages
  4071--4081. Proceedings of Machine Learning Research (PMLR), 2020.

\bibitem[He et~al.(2016)He, Zhang, Ren, and Sun]{he2016deep}
K.~He, X.~Zhang, S.~Ren, and J.~Sun.
\newblock Deep residual learning for image recognition.
\newblock In \emph{Conference on Computer Vision and Pattern Recognition
  (CVPR)}, pages 770--778, 2016.

\bibitem[Hendrycks et~al.(2018)Hendrycks, Mazeika, Wilson, and
  Gimpel]{hendrycks2018using}
D.~Hendrycks, M.~Mazeika, D.~Wilson, and K.~Gimpel.
\newblock Using trusted data to train deep networks on labels corrupted by
  severe noise.
\newblock In \emph{Conference on Neural Information Processing Systems
  (NeurIPS)}, 2018.

\bibitem[Hooker et~al.(2019)Hooker, Courville, Dauphin, and
  Frome]{hooker2019selective}
S.~Hooker, A.~Courville, Y.~Dauphin, and A.~Frome.
\newblock Selective brain damage: Measuring the disparate impact of model
  pruning.
\newblock \emph{arXiv preprint arXiv:1911.05248}, 2019.

\bibitem[Huang et~al.(2019)Huang, Emam, Goldblum, Fowl, Terry, Huang, and
  Goldstein]{huang2019understanding_generalization}
W.~R. Huang, Z.~Emam, M.~Goldblum, L.~Fowl, J.~K. Terry, F.~Huang, and
  T.~Goldstein.
\newblock Understanding generalization through visualizations.
\newblock \emph{arXiv preprint arXiv:1906.03291}, 2019.

\bibitem[Jiang et~al.(2020)Jiang, Huang, Liu, and
  Yang]{pmlr-v119-jiang20c-synthetic-noise}
L.~Jiang, D.~Huang, M.~Liu, and W.~Yang.
\newblock Beyond synthetic noise: Deep learning on controlled noisy labels.
\newblock In \emph{International Conference on Machine Learning (ICML)}, volume
  119 of \emph{Proceedings of Machine Learning Research (PMLR)}, pages
  4804--4815. Proceedings of Machine Learning Research (PMLR), 13--18 Jul 2020.
\newblock URL \url{http://proceedings.mlr.press/v119/jiang20c.html}.

\bibitem[Jindal et~al.(2016)Jindal, Nokleby, and
  Chen]{7837934_multiclass_learning_with_noise_using_dropout}
I.~Jindal, M.~Nokleby, and X.~Chen.
\newblock Learning deep networks from noisy labels with dropout regularization.
\newblock In \emph{International Conference on Data Mining (ICDM)}, pages
  967--972, Dec. 2016.
\newblock \doi{10.1109/ICDM.2016.0121}.

\bibitem[Kremer et~al.(2018)Kremer, Sha, and
  Igel]{pmlr-robust-active-label-correction-2018}
J.~Kremer, F.~Sha, and C.~Igel.
\newblock Robust active label correction.
\newblock In \emph{Proceedings of Machine Learning Research (PMLR)}, volume~84
  of \emph{Proceedings of Machine Learning Research}, pages 308--316, Playa
  Blanca, Lanzarote, Canary Islands, 09--11 Apr 2018. Proceedings of Machine
  Learning Research (PMLR).
\newblock URL \url{http://proceedings.mlr.press/v84/kremer18a.html}.

\bibitem[Krizhevsky and Hinton(2009)]{cifar10}
A.~Krizhevsky and G.~Hinton.
\newblock Learning multiple layers of features from tiny images.
\newblock \emph{Master's thesis, Department of Computer Science, University of
  Toronto}, 2009.
\newblock URL \url{http://www.cs.toronto.edu/~kriz/cifar.html}.

\bibitem[Lecun et~al.(1998)Lecun, Bottou, Bengio, and Haffner]{mnist}
Y.~Lecun, L.~Bottou, Y.~Bengio, and P.~Haffner.
\newblock Gradient-based learning applied to document recognition.
\newblock In \emph{Proceedings of the IEEE}, pages 2278--2324, 1998.

\bibitem[Li et~al.(2021)Li, Rao, Blase, Zhang, Chu, and Zhang]{li2021cleanml}
P.~Li, X.~Rao, J.~Blase, Y.~Zhang, X.~Chu, and C.~Zhang.
\newblock {CleanML:} a study for evaluating the impact of data cleaning on ml
  classification tasks.
\newblock In \emph{IEEE International Conference on Data Engineering}, 2021.

\bibitem[Li et~al.(2017)Li, Wang, Li, Agustsson, and Van~Gool]{li2017webvision}
W.~Li, L.~Wang, W.~Li, E.~Agustsson, and L.~Van~Gool.
\newblock Webvision database: Visual learning and understanding from web data.
\newblock \emph{arXiv preprint arXiv:1708.02862}, 2017.

\bibitem[Lipton et~al.(2018)Lipton, Wang, and
  Smola]{icml_lipton_label_shift_confusion_matrix}
Z.~Lipton, Y.-X. Wang, and A.~Smola.
\newblock Detecting and correcting for label shift with black box predictors.
\newblock In \emph{International Conference on Machine Learning (ICML)}, 2018.

\bibitem[{List of Datasets for Machine Learning
  Research}(2018)]{machine-learning-datasets-wikipedia-citations}
{List of Datasets for Machine Learning Research}.
\newblock List of datasets for machine learning research --- {W}ikipedia{,} the
  free encyclopedia.
\newblock
  \url{https://en.wikipedia.org/wiki/List_of_datasets_for_machine-learning_research},
  2018.
\newblock [Online; accessed 22-October-2018].

\bibitem[Maas et~al.(2011)Maas, Daly, Pham, Huang, Ng, and Potts]{imdb}
A.~L. Maas, R.~E. Daly, P.~T. Pham, D.~Huang, A.~Y. Ng, and C.~Potts.
\newblock Learning word vectors for sentiment analysis.
\newblock In \emph{Annual Conference of the Association for Computational
  Linguistics (ACL)}, pages 142--150, Portland, Oregon, USA, June 2011. Annual
  Conference of the Association for Computational Linguistics (ACL).
\newblock URL \url{http://www.aclweb.org/anthology/P11-1015}.

\bibitem[Mahajan et~al.(2018)Mahajan, Girshick, Ramanathan, He, Paluri, Li,
  Bharambe, and Van
  Der~Maaten]{fair_laurens_van_der_maaten_limits_weak_supervision_2018}
D.~Mahajan, R.~Girshick, V.~Ramanathan, K.~He, M.~Paluri, Y.~Li, A.~Bharambe,
  and L.~Van Der~Maaten.
\newblock Exploring the limits of weakly supervised pretraining.
\newblock \emph{European Conference on Computer Vision (ECCV)}, pages 181--196,
  2018.

\bibitem[McAuley et~al.(2015)McAuley, Targett, Shi, and van~den
  Hengel]{amazon-reviews}
J.~McAuley, C.~Targett, Q.~Shi, and A.~van~den Hengel.
\newblock Image-based recommendations on styles and substitutes.
\newblock In \emph{Special Interest Group on Information Retrieval (SIGIR)},
  pages 43--52. ACM, 2015.
\newblock ISBN 978-1-4503-3621-5.
\newblock \doi{10.1145/2766462.2767755}.
\newblock URL \url{http://doi.acm.org/10.1145/2766462.2767755}.

\bibitem[Mitchell(1999)]{20newsgroups}
T.~Mitchell.
\newblock Twenty newsgroups dataset.
\newblock \url{https://archive.ics.uci.edu/ml/datasets/Twenty+Newsgroups},
  1999.

\bibitem[Natarajan et~al.(2013)Natarajan, Dhillon, Ravikumar, and
  Tewari]{NIPS2013_5073}
N.~Natarajan, I.~S. Dhillon, P.~K. Ravikumar, and A.~Tewari.
\newblock Learning with noisy labels.
\newblock In \emph{Conference on Neural Information Processing Systems
  (NeurIPS)}, pages 1196--1204, 2013.
\newblock URL
  \url{http://papers.nips.cc/paper/5073-learning-with-noisy-labels.pdf}.

\bibitem[Northcutt et~al.(2017)Northcutt, Wu, and
  Chuang]{northcutt2017rankpruning}
C.~G. Northcutt, T.~Wu, and I.~L. Chuang.
\newblock Learning with confident examples: Rank pruning for robust
  classification with noisy labels.
\newblock In \emph{Conference on Uncertainty in Artificial Intelligence (UAI)},
  2017.

\bibitem[Northcutt et~al.(2021)Northcutt, Jiang, and
  Chuang]{northcutt2021confidentlearning}
C.~G. Northcutt, L.~Jiang, and I.~Chuang.
\newblock Confident learning: Estimating uncertainty in dataset labels.
\newblock \emph{Journal of Artificial Intelligence Research}, 70:\penalty0
  1373--1411, 2021.

\bibitem[Patrini et~al.(2016)Patrini, Nielsen, Nock, and
  Carioni]{DBLP:conf/icml/PatriniNNC16}
G.~Patrini, F.~Nielsen, R.~Nock, and M.~Carioni.
\newblock Loss factorization, weakly supervised learning and label noise
  robustness.
\newblock In \emph{International Conference on Machine Learning (ICML)}, pages
  708--717, 2016.

\bibitem[Patrini et~al.(2017)Patrini, Rozza, Krishna~Menon, Nock, and
  Qu]{patrini2017making}
G.~Patrini, A.~Rozza, A.~Krishna~Menon, R.~Nock, and L.~Qu.
\newblock Making deep neural networks robust to label noise: A loss correction
  approach.
\newblock In \emph{Conference on Computer Vision and Pattern Recognition
  (CVPR)}, 2017.

\bibitem[Ratner et~al.(2016)Ratner, De~Sa, Wu, Selsam, and
  R\'{e}]{ratner_chris_re_large_training_sets_NIPS2016_6523}
A.~J. Ratner, C.~M. De~Sa, S.~Wu, D.~Selsam, and C.~R\'{e}.
\newblock Data programming: Creating large training sets, quickly.
\newblock In \emph{Conference on Neural Information Processing Systems
  (NeurIPS)}, pages 3567--3575, 2016.
\newblock URL
  \url{http://papers.nips.cc/paper/6523-data-programming-creating-large-training-sets-quickly.pdf}.

\bibitem[Recht et~al.(2019)Recht, Roelofs, Schmidt, and
  Shankar]{recht2019imagenet}
B.~Recht, R.~Roelofs, L.~Schmidt, and V.~Shankar.
\newblock Do imagenet classifiers generalize to imagenet?
\newblock In \emph{International Conference on Machine Learning (ICML)}, pages
  5389--5400, 2019.

\bibitem[Rolnick et~al.(2017)Rolnick, Veit, Belongie, and
  Shavit]{rolnick2017deep}
D.~Rolnick, A.~Veit, S.~Belongie, and N.~Shavit.
\newblock Deep learning is robust to massive label noise.
\newblock \emph{arXiv preprint arXiv:1705.10694}, 2017.

\bibitem[Sambasivan et~al.(2021)Sambasivan, Kapania, Highfill, Akrong,
  Paritosh, and Aroyo]{chi2021data}
N.~Sambasivan, S.~Kapania, H.~Highfill, D.~Akrong, P.~Paritosh, and L.~M.
  Aroyo.
\newblock "{E}veryone wants to do the model work, not the data work": Data
  cascades in high-stakes ai.
\newblock In \emph{Human Factors in Computing Systems (CHI)}, 2021.

\bibitem[Shankar et~al.(2020)Shankar, Roelofs, Mania, Fang, Recht, and
  Schmidt]{imageneterror2020shankar}
V.~Shankar, R.~Roelofs, H.~Mania, A.~Fang, B.~Recht, and L.~Schmidt.
\newblock Evaluating machine accuracy on {I}mage{N}et.
\newblock In \emph{International Conference on Machine Learning (ICML)}, volume
  119 of \emph{Proceedings of Machine Learning Research}, pages 8634--8644.
  Proceedings of Machine Learning Research (PMLR), 13--18 Jul 2020.

\bibitem[Simonyan and Zisserman(2014)]{simonyan2014vgg}
K.~Simonyan and A.~Zisserman.
\newblock Very deep convolutional networks for large-scale image recognition.
\newblock \emph{arXiv preprint arXiv:1409.1556}, 2014.

\bibitem[Sukhbaatar et~al.(2015)Sukhbaatar, Bruna, Paluri, Bourdev, and
  Fergus]{Sukhbaatar_fergus_iclr_2015}
S.~Sukhbaatar, J.~Bruna, M.~Paluri, L.~Bourdev, and R.~Fergus.
\newblock Training convolutional networks with noisy labels.
\newblock In \emph{International Conference on Learning Representations
  (ICLR)}, pages 1--11, 2015.
\newblock ISBN 9781611970685.
\newblock \doi{10.1051/0004-6361/201527329}.
\newblock URL \url{http://arxiv.org/abs/1406.2080}.

\bibitem[Sun et~al.(2017)Sun, Shrivastava, Singh, and
  Gupta]{gupta_unreasonable_effectiveness_of_data_2017}
C.~Sun, A.~Shrivastava, S.~Singh, and A.~Gupta.
\newblock Revisiting unreasonable effectiveness of data in deep learning era.
\newblock In \emph{International Conference on Computer Vision (ICCV)}, Oct
  2017.

\bibitem[Tsipras et~al.(2020)Tsipras, Santurkar, Engstrom, Ilyas, and
  Madry]{tsipras2020imagenet}
D.~Tsipras, S.~Santurkar, L.~Engstrom, A.~Ilyas, and A.~Madry.
\newblock From imagenet to image classification: Contextualizing progress on
  benchmarks.
\newblock In \emph{International Conference on Machine Learning}, pages
  9625--9635. Proceedings of Machine Learning Research (PMLR), 2020.

\bibitem[Van~Rooyen et~al.(2015)Van~Rooyen, Menon, and
  Williamson]{rooyen_menon_unhinged_nips15}
B.~Van~Rooyen, A.~Menon, and R.~C. Williamson.
\newblock Learning with symmetric label noise: The importance of being
  unhinged.
\newblock In \emph{Conference on Neural Information Processing Systems
  (NeurIPS)}, pages 10--18, 2015.
\newblock URL
  \url{http://papers.nips.cc/paper/5941-learning-with-symmetric-label-noise-the-importance-of-being-unhinged}.

\bibitem[Wang et~al.(2021)Wang, Liu, and Levy]{wang2021fair}
J.~Wang, Y.~Liu, and C.~Levy.
\newblock Fair classification with group-dependent label noise.
\newblock In \emph{Proceedings of the ACM Conference on Fairness,
  Accountability, and Transparency}, 2021.

\bibitem[Wei et~al.(2018)Wei, Lee, Liu, and Ma]{wei2018nomralizedmaxmargin}
C.~Wei, J.~D. Lee, Q.~Liu, and T.~Ma.
\newblock On the margin theory of feedforward neural networks.
\newblock \emph{Computing Research Repository (CoRR)}, 2018.
\newblock URL \url{http://arxiv.org/abs/1810.05369}.

\bibitem[Xu et~al.(2019)Xu, Cao, Kong, and
  Wang]{neurips2019novelinformationtheory}
Y.~Xu, P.~Cao, Y.~Kong, and Y.~Wang.
\newblock L\_dmi: A novel information-theoretic loss function for training deep
  nets robust to label noise.
\newblock In \emph{Conference on Neural Information Processing Systems
  (NeurIPS)}, pages 6225--6236, 2019.

\bibitem[Zhang et~al.(2017)Zhang, Sheng, Li, and Wu]{zhang2017improving}
J.~Zhang, V.~S. Sheng, T.~Li, and X.~Wu.
\newblock Improving crowdsourced label quality using noise correction.
\newblock \emph{IEEE Transactions on Neural Networks and Learning Systems},
  29\penalty0 (5):\penalty0 1675--1688, 2017.

\bibitem[Zoph et~al.(2018)Zoph, Vasudevan, Shlens, and Le]{zoph2018learning}
B.~Zoph, V.~Vasudevan, J.~Shlens, and Q.~V. Le.
\newblock Learning transferable architectures for scalable image recognition.
\newblock In \emph{Conference on Computer Vision and Pattern Recognition
  (CVPR)}, pages 8697--8710, 2018.

\end{thebibliography}
\interlinepenalty=10000
\bibliographystyle{abbrvnat}

\beginsupplement
\clearpage
\appendix

\begin{center}{
    \LARGE \textbf{Appendix: \ \papertitle}
    }
\end{center}

\section{Datasets} \label{app:datasets}

For our study, we select 10 of the most-cited, open-source datasets created in the last 20 years from the \href{https://en.wikipedia.org/wiki/List_of_datasets_for_machine-learning_research}{Wikipedia List of ML Research Datasets} \citep{machine-learning-datasets-wikipedia-citations}, with preference for diversity across computer vision, NLP, sentiment analysis, and audio modalities. Citation counts were obtained via the Microsoft Cognitive API\@. In total, we evaluate six visual datasets: MNIST, CIFAR-10, CIFAR-100, Caltech-256, ImageNet, and QuickDraw; three text datasets: 20news, IMDB, and Amazon Reviews; and one audio dataset: AudioSet.

\subsection{Dataset details} 

For each of the datasets we investigate, we summarize the original data collection and labeling procedure as they pertain to potential label errors.

\textbf{MNIST \citep{mnist}.} MNIST is a database of binary images of handwritten digits. The dataset was constructed from Handwriting Sample Forms distributed to Census Bureau employees and high school students; the ground-truth labels were determined by matching digits to the instructions of the task to copy a particular set of digits \citep{grother1995nist_mnistcuration}. Label errors may arise from failure to follow instructions or from handwriting ambiguities.

\textbf{CIFAR-10 / CIFAR-100 \citep{cifar10}.} The CIFAR-10 and CIFAR-100 datasets are collections of small $32 \times 32$ images and labels from a set of 10 or 100 classes, respectively. The images were collected by searching the internet for the class label. Human labelers were instructed to select images that matched their class label (query term) by filtering out mislabeled images. Images were intended to only have one prominent instance of the object, but could be partially occluded as long as it was identifiable to the labeler.

\textbf{Caltech-256 \citep{caltech}.} Caltech-256 is a database of images sorted into 256 classes, plus an extra class called ``clutter''. Images were scraped from image search engines. Four human labelers were instructed to rate the images into ``good,'' ``bad,'' and ``not applicable,'' eliminating the images that were confusing, occluded, cluttered, artistic, or not an example of the object category from the dataset.
Because no explicit test set is provided, we study label errors in the entire dataset to ensure coverage of any test set split used by practitioners.
\textbf{Modifications}: In our study, we ignore data with the ambiguous ``clutter'' label (class 257) and consider only the images labeled class 1 to class 256.

\textbf{ImageNet \citep{imagenet}.} ImageNet is a database of images belonging to one of 1,000 classes. Images were scraped by querying words from WordNet ``synonym sets'' (synsets) on several image search engines. The images were labeled by Amazon Mechanical Turk workers who were asked whether each image contains objects of a particular given synset. Workers were instructed to select images that contain objects of a given subset regardless of occlusions, number of objects, and clutter to ``ensure diversity'' in the dataset's images.

\textbf{QuickDraw \citep{quickdraw2017google}.} The Quick, Draw! dataset contains more than 1 billion doodles collected from users of an experimental game to benchmark image classification models. Users were instructed to draw pictures corresponding to a given label, but the drawings may be ``incomplete or may not match the label.'' Because no explicit test set is provided, we study label errors in the entire dataset to ensure coverage of any test set split used by practitioners.

\textbf{20news \citep{20newsgroups}.} The 20 Newsgroups dataset is a collection of articles posted to Usenet newsgroups used to benchmark text classification and clustering models. The label for each example is the newsgroup it was originally posted in (e.g. ``misc.forsale''), so it is obtained during the overall data collection procedure.

\textbf{IMDB \citep{imdb}.} The IMDB Large Movie Review Dataset is a collection of movie reviews to benchmark binary sentiment classification. The labels were determined by the user's review: a score $\leq 4$ out of $10$ is considered negative; $\geq 7$ out of $10$ is considered positive.

\textbf{Amazon Reviews \citep{amazon-reviews}.} The Amazon Reviews dataset is a collection of textual reviews and 5-star ratings from Amazon customers used to benchmark sentiment analysis models. We use the 5-core (9.9 GB) variant of the dataset. \textbf{Modifications}: In our study, 2-star and 4-star reviews are removed due to ambiguity with 1-star and 5-star reviews, respectively. If these reviews were left in the dataset, they could inflate error counts. Because no explicit test set is provided, we study label errors in the entire dataset to ensure coverage of any test set split used by practitioners.

\textbf{AudioSet \citep{audioset}.} AudioSet is a collection of 10-second sound clips drawn from YouTube videos and multiple labels describing the sounds that are present in the clip. Three human labelers independently rated the presence of one or more labels (as ``present,'' ``not present,'' and ``unsure''), and majority agreement was required to assign a label. The authors note that spot checking revealed some label errors due to ``confusing labels, human error, and difference in detection of faint/non-salient audio events.''

\clearpage
\section{Mechanical Turk details} \label{sec:mturk-interface}

\paragraph{Mechanical Turk budget} 

Mechanical Turk workers were paid an hourly rate of \$7.20 (based on an estimated evaluation time of 5 seconds per image). In total, we spent \$1623.29 on human verification experiments on Mechanical Turk. Results would likely improve with a larger budget.

\begin{figure*}[!h]
    \centering
    \includegraphics[width=.99\linewidth]{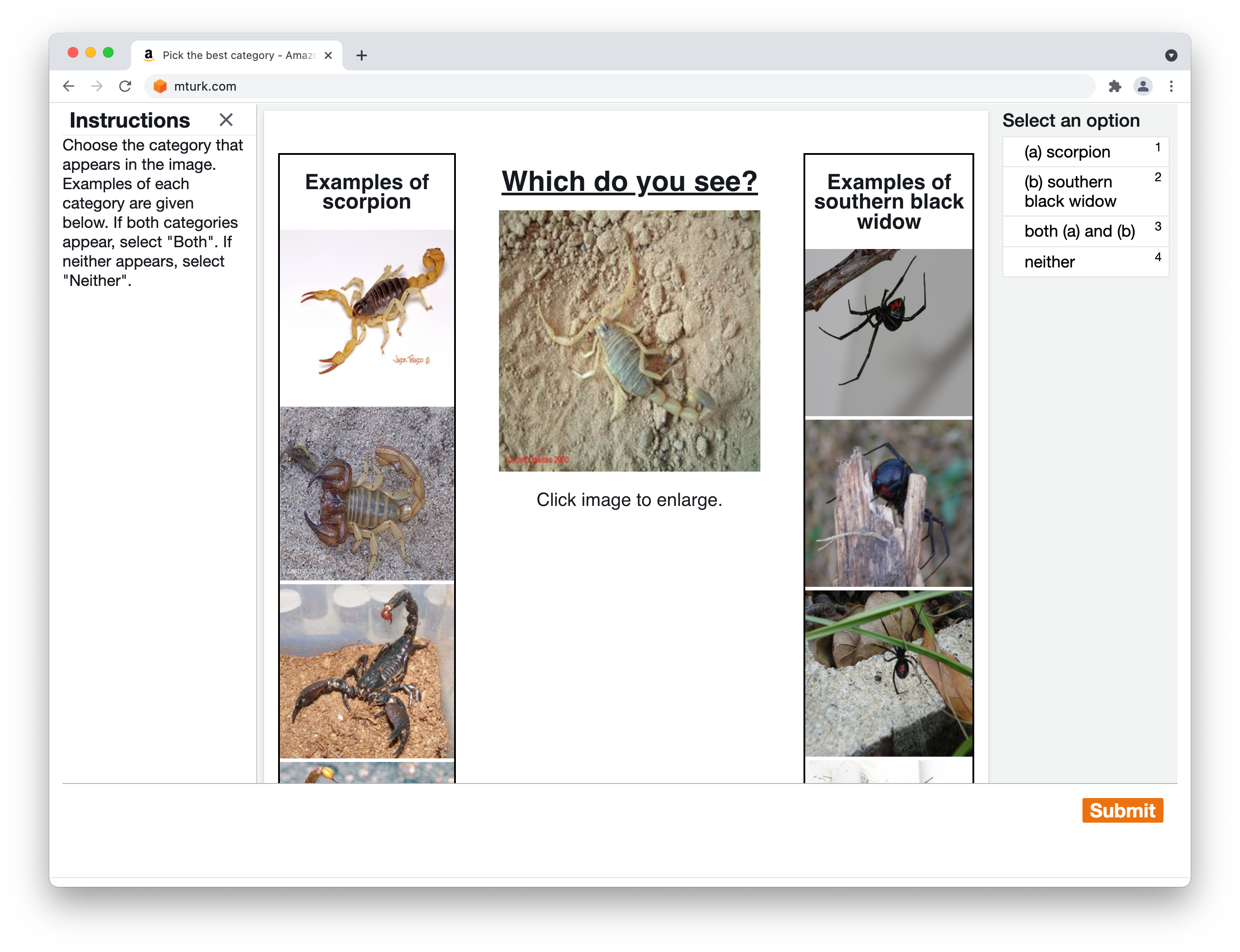}
    \caption{Mechanical Turk worker interface showing an example from ImageNet (with given label ``southern black window''). For each data point algorithmically identified as a potential label error, the interface presents the data point, along with examples belonging to the given class. The interface also shows data points belonging to the confidently predicted class (in this case, ``scorpion''). Either the given label is shown as option (a) and the predicted label is shown as option (b), or vice versa (chosen randomly). The worker is asked whether the image belongs to class (a), (b), both, or neither.}
    \label{fig:mturk-interface}
\end{figure*}

\begin{figure*}[!h]
    \centering
    \includegraphics[width=.3\linewidth]{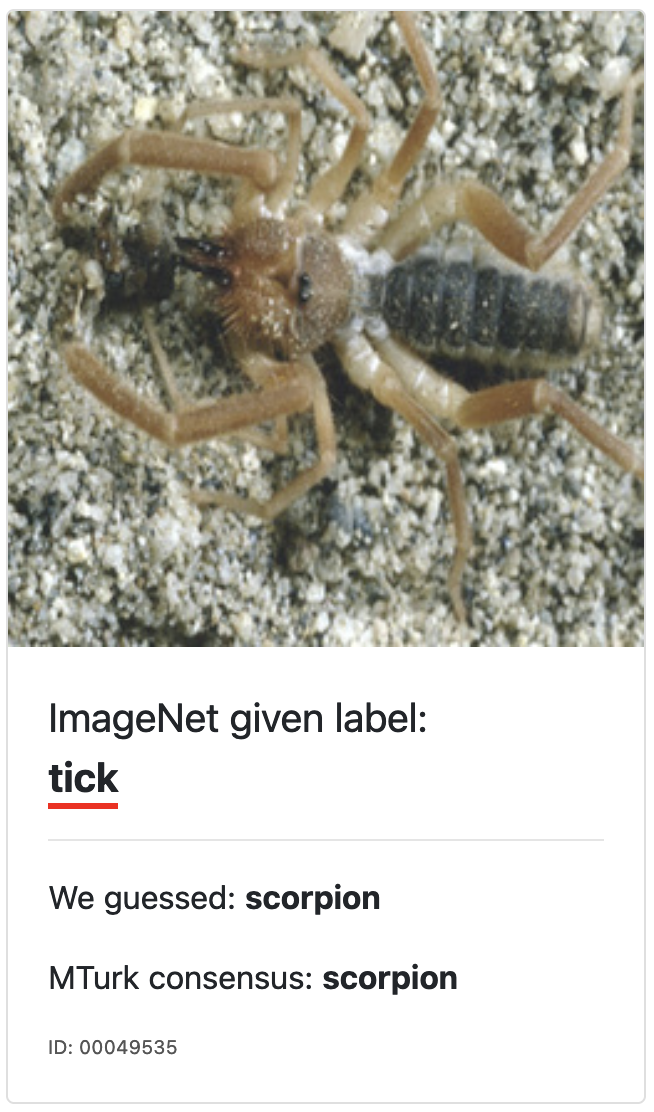}
    \caption{An example from \url{https://labelerrors.com} that Mechanical Turk workers got wrong. The image clearly doesn't match the ImageNet given label ``tick,'' but upon close inspection, it does not match the predicted label ``scorpion'' either. The insect shown is in fact an arachnid of the order Solifugae, commonly known as camel spiders or wind scorpions. Despite the common name, this animal is not a true scorpion.}
    \label{fig:mturk-mistake}
\end{figure*}

\clearpage
\section{Details of confident learning (CL) for finding label errors} \label{confidentlearning}

Here we summarize CL joint estimation and how it is used to algorithmically flag candidates with likely label errors for subsequent human review. 
An unnormalized representation of the joint distribution between observed and true label, called the \emph{confident joint} and denoted $\cj$, is estimated by counting all the examples with noisy label $\tilde{y}=i$, with high probability of actually belonging to label $y^*=j$. This binning can be expressed as:

\begin{equation} \label{eqn:cj}
    \cj = \lvert \{\bm{x} \in \bm{X}_{\tilde{y} = i} : \,\, \hat{p} (\tilde{y} = j ;\bm{x}, \bm{\theta})  \ge t_j \} \rvert \nonumber
\end{equation}

where $\bm{x}$ is a data example (e.g. an image), $\bm{X}_{\tilde{y} = i}$ is the set of examples with noisy label $\tilde{y} = i$, $\hat{p} (\tilde{y} = j ;\bm{x}, \bm{\theta})$ is the out-of-sample predicted probability that example $\bm{x}$ actually belongs to noisy class $\tilde{y} = j$ (even though its given label $\tilde{y} = i$) for a given model $\bm{\theta}$. Finally, $t_j$ is a per-class threshold that, in comparison to other confusion matrix approaches, provides robustness to heterogeneity in class distributions and class distributions, defined as:

\begin{equation} \label{eqn:cj_threshold}
     t_j = \frac{1}{|\bm{X}_{\tilde{y}=j}|} \sum_{\bm{x} \in \bm{X}_{\tilde{y}=j}} \hat{p}(\tilde{y}=j; \bm{x}, \bm{\theta})
\end{equation}

A caveat occurs when an example is confidently counted into more than one bin. When this occurs, the example is only counted in the $ \argmax_{l \in [m]} \hat{p} (\tilde{y} = l ;\bm{x}, \bm{\theta})$ bin.

$\joint$ is estimated by normalizing $\cj$, as follows:
\begin{equation} \label{eqn_calibration}
    \estjointlong = \frac{\frac{\bm{C}_{\tilde{y}=i, y^*=j}}{\sum_{j \in [m]} \bm{C}_{\tilde{y}=i, y^*=j}} \cdot \lvert \bm{X}_{\tilde{y}=i} \rvert}{\sum\limits_{i \in [m], j \in [m]} \left( \frac{\bm{C}_{\tilde{y}=i, y^*=j}}{\sum_{j \in [m]} \bm{C}_{\tilde{y}=i, y^*=j}} \cdot \lvert \bm{X}_{\tilde{y}=i} \rvert \right)}
\end{equation}
The numerator calibrates $\sum_j \estjointlong = \lvert \bm{X}_i \rvert / \sum_{i \in [m]}  \lvert \bm{X}_i \rvert, \forall i \smallin [m]$ so that row-sums match the observed prior over noisy labels. The denominator makes the distribution sum to 1.

\clearpage
\section{Failure modes of confident learning} \label{sec:failure_modes_math}

Confident learning can fail to exactly estimate $\partition$ (the set of examples with noisy label $i$ and actual label $j$) when either:

\begin{itemize}[topsep=0pt]
    \item \textbf{Case 1}: $\hat{p}(\tilde{y} \smalleq j; \bm{x}, \bm{\theta}) < t_j \longrightarrow \bm{x} \not\in \estpartition$, \; or 
    \item \textbf{Case 2}:  $\hat{p}(\tilde{y} \smalleq k; \bm{x}, \bm{\theta}) \geq t_k \longrightarrow \bm{x} \in \hat{\bm{X}}_{ \tilde{y} \smalleq i,y^* \smalleq k}$, \; for some $k \neq j$
\end{itemize}

\noindent where $t_j$ is the per-class average threshold (Eqn. \ref{eqn:cj_threshold} above, in Appendix~\ref{confidentlearning}). In the real-world datasets we study, the predicted probabilities are noisy such that $\predprobshortj = \perfprobshortj + \errorxj$, where $\predprobshortj$ is shorthand for $\hat{p}(\tilde{y} \smalleq j; \bm{x}, \bm{\theta})$; $\perfprobshortj$ is the ideal/non-noisy predicted probability; and $\errorxj \in \mathcal{R}$ is the error/deviation from ideal. Unlike learning with perfect labels, $\perfprobshortj$ is not always 0 or 1 because in our setting some classes are mislabeled as other classes some fraction of the time. Expressing the two failure cases in terms of error, we have:

\begin{itemize}[topsep=0pt]
    \item \textbf{Case 1}: $\errorxj < t_j - \perfprobshortj \longrightarrow \bm{x} \not\in \estpartition$, or
    \item \textbf{Case 2}: $\errorxk \geq t_k - \perfprobshortk \longrightarrow \bm{x} \in \hat{\bm{X}}_{ \tilde{y} \smalleq i,y^* \smalleq k}$, for some $k \neq j$
\end{itemize}

\noindent Case 1 bounds the error of $\hat{p}(\tilde{y} \smalleq j; \bm{x}, \bm{\theta})$ (in the limit to $-\infty$) and Case 2 bound the error of $\hat{p}(\tilde{y} \smalleq j; \bm{x}, \bm{\theta})$ (in the limit to $\infty$) such that when either occurs, $\exists (i, j) \small{\in} [m] \small{\times} [m]$, s.t. $\estpartition \neq \partition$, i.e., we imperfectly estimate the label errors prior to human validation. Figure \ref{fig:failure_mode} shows uniquely challenging examples (with excessively erroneous $\hat{p}(\tilde{y} \smalleq j; \bm{x}, \bm{\theta})$) when these failure mode cases potentially occur.

\clearpage
\section{Reproducibility and computational requirements} \label{sec:reproduce_computation}
For all 10 datasets, label errors were found using a Linux 18.04 LTS server comprising 128GB of memory, an Intel Core i9-9820X Skylake X 10-Core 3.3GHz, and one RTX 2080 TI GPU. We open-source a single script to reproduce the label errors for every dataset at \url{https://github.com/cleanlab/label-errors/blob/main/examples/Tutorial\%20-\%20How\%20To\%20Find\%20Label\%20Errors\%20With\%20CleanLab.ipynb}. Reproducing the label errors for all 10 datasets using this tutorial takes about 5 minutes on a modern consumer-grade laptop (e.g., a 2021 Apple M1 MacBook Air).

\clearpage
\section{Additional findings on implications of label errors in test data}
\begin{figure*}[!ht]
\begin{subfigure}{0.5\textwidth}
\includegraphics[width=0.9\linewidth, ]{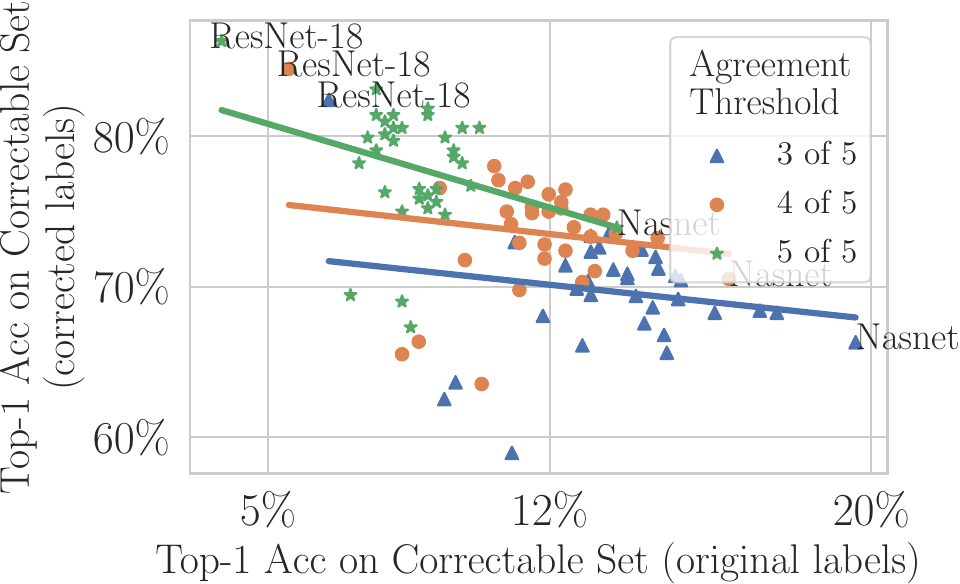} 
\vskip -0.05in
\caption{Top-1 Accuracy.}
\label{fig:orig_vs_corrected_all_top5}
\end{subfigure}
\begin{subfigure}{0.5\textwidth}
\includegraphics[width=0.9\linewidth, ]{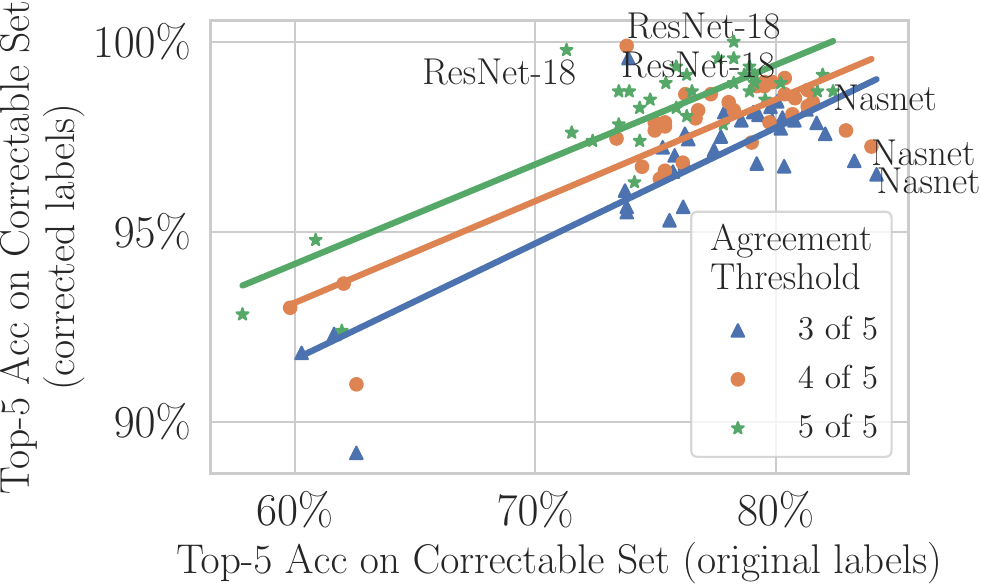}
\vskip -0.05in
\caption{Top-5 Accuracy.}
\label{fig:orig_vs_corrected_all_top1}
\end{subfigure}

\caption{Benchmark ranking comparison of 34 pre-trained models on the ImageNet val set (used as test data here) for various settings of the agreement threshold. Top-5 benchmarks are  unchanged by removing label errors (a), but change drastically on the correctable subset with original (erroneous) labels versus corrected labels. Corrected test set sizes: 1428 ($\blacktriangle$), 960 ($\bullet$), 468 ($\star$).}
\label{fig:orig_vs_corrected_all}
\end{figure*}

Here we provide some additional details/results to complement Section \ref{sec:implications_discussion} from the main text. 
Figure \ref{fig:imagenet_benchmarking} depicts how the benchmarking rankings on the correctable subset of ImageNet examples change significantly for an \emph{agreement threshold} $= 5$, meaning $5$ of $5$ human raters need to independently select the same alternative label for that data point and a new label to be included in the accuracy evaluation. To ascertain that the results of this figure are not due to the setting of the agreement threshold, the results for all three settings of the agreement threshold are shown in Sub-figure \ref{fig:orig_vs_corrected_all_top1}. Observe the negative correlation (for top-1 accuracy) occurs in all three settings. Furthermore, observe that this negative correlation no longer holds when top-5 accuracy is used (shown in \ref{fig:orig_vs_corrected_all_top5}), likely because many of these models use a loss which maximizes (and overfits to noise) based on top-1 accuracy, not top-5 accuracy. Regardless of whether top-1 or top-5 accuracy is used, model benchmark rankings change significantly on the correctable set in comparison to the original test set (see Table \ref{tab:imagenet_benchmarking_table}).

\begin{figure*}[!ht]
\centering
\hspace*{.2in}\includegraphics[width=0.88\linewidth, ]{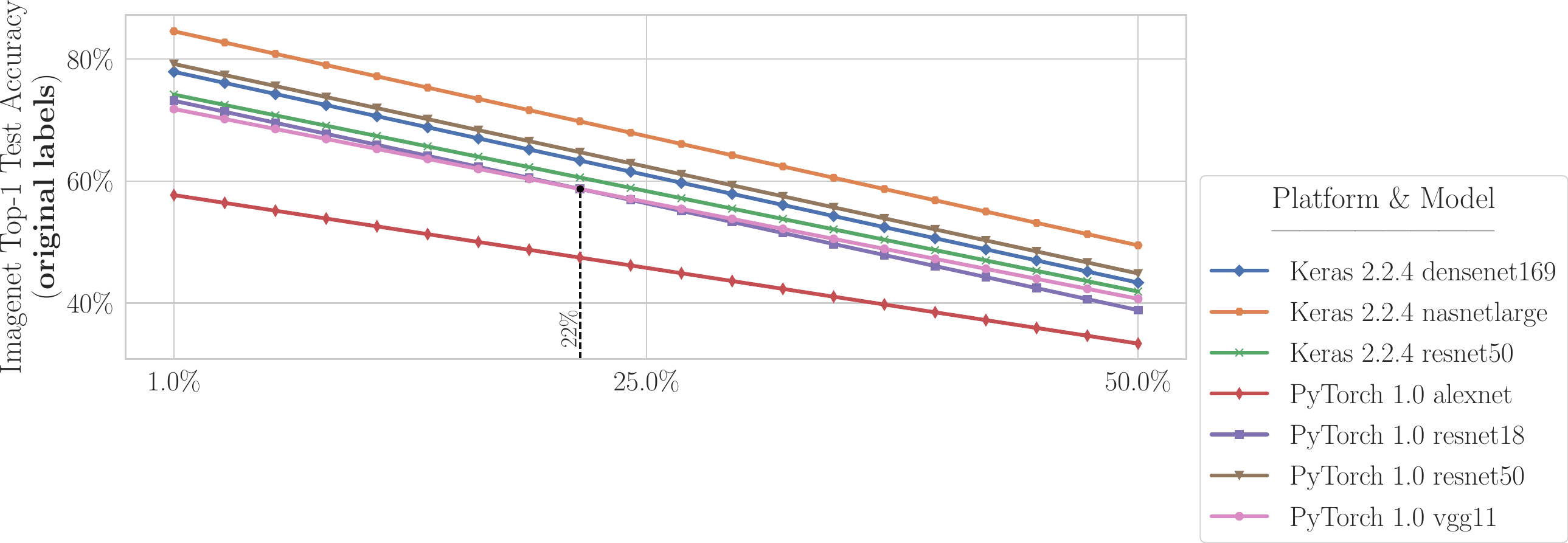}
\vskip -0.45in
\hspace*{.2in}\hspace*{-1.18in}\includegraphics[width=0.666\linewidth, ]{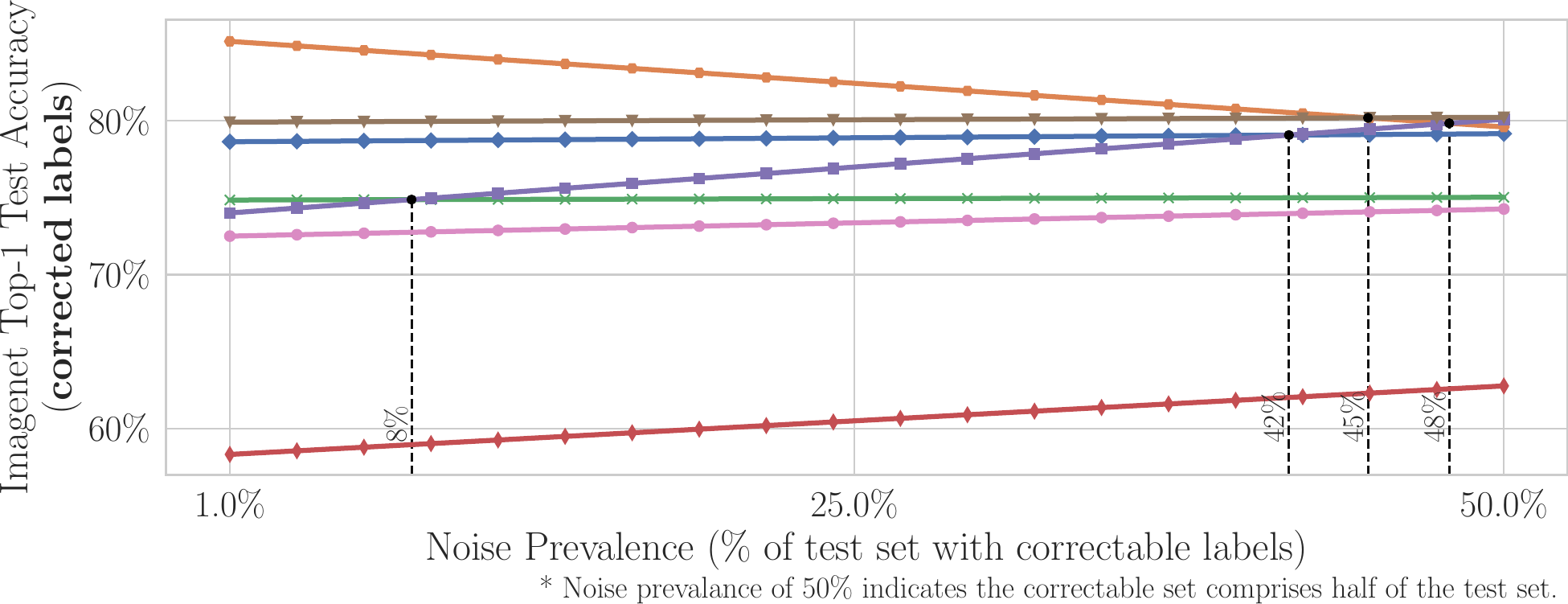}
\caption{ImageNet top-1 original accuracy (top panel) and top-1 corrected accuracy (bottom panel) vs Noise Prevalence with agreement threshold = 5 (instead of threshold = 3, c.f., Figure \ref{fig:imagenet_original_noise_prevalance_agreement_threshold_3}).
}
\label{fig:imagenet_original_noise_prevalance_agreement_threshold_5}
\end{figure*}

\begin{table*}[ht]
\setlength\tabcolsep{2pt} % Makes table columns tighter
    \caption{Individual accuracy scores for Sub-figure \ref{fig:orig_vs_corrected} with \emph{agreement threshold = 3 of 5}. Acc@1 stands for the (top-1 validation) original accuracy on the correctable set, in terms of original ImageNet examples and labels. \emph{cAcc@1} stands for the (top-1 validation) corrected accuracy on the correctable set of ImageNet examples with correct labels. To be corrected, at least 3 of 5 Mechanical Turk raters had to independently agree on a new label, proposed by us using the class with the $\argmax$ probability for the example.}
    \label{tab:imagenet_benchmarking_table}
    \center
    \resizebox{.9\linewidth}{!}{ %Completely zooms in or zooms out (shrinks) entire table!
        \begin{tabular}{llrrrrrrrr}
        \toprule
            Platform &              Model &  Acc@1 &  cAcc@1 &  Acc@5 &  cAcc@5 &  Rank@1 &  cRank@1 &  Rank@5 &  cRank@5 \\
        \midrule
         PyTorch 1.0 &           resnet18 &   6.51 &   82.42 &  73.81 &   99.58 &      34 &        1 &      30 &        1 \\
         PyTorch 1.0 &           resnet50 &  13.52 &   73.74 &  79.97 &   98.46 &      20 &        2 &      11 &        2 \\
         PyTorch 1.0 &           vgg19\_bn &  13.03 &   73.39 &  79.97 &   97.97 &      23 &        3 &      10 &        9 \\
         PyTorch 1.0 &           vgg11\_bn &  11.13 &   72.97 &  76.26 &   97.55 &      30 &        4 &      22 &       15 \\
         PyTorch 1.0 &           resnet34 &  13.24 &   72.62 &  77.80 &   98.11 &      21 &        5 &      18 &        6 \\
         PyTorch 1.0 &        densenet169 &  14.15 &   72.55 &  79.62 &   98.32 &      16 &        6 &      12 &        3 \\
         PyTorch 1.0 &        densenet121 &  14.29 &   72.48 &  78.64 &   97.97 &      14 &        7 &      16 &       11 \\
         PyTorch 1.0 &              vgg19 &  13.03 &   72.34 &  79.34 &   98.04 &      22 &        8 &      13 &        8 \\
         PyTorch 1.0 &          resnet101 &  14.64 &   71.99 &  81.16 &   98.25 &      11 &        9 &       5 &        4 \\
         PyTorch 1.0 &              vgg16 &  12.39 &   71.43 &  77.52 &   97.20 &      28 &       10 &      19 &       19 \\
         PyTorch 1.0 &        densenet201 &  14.71 &   71.22 &  80.81 &   97.97 &      10 &       11 &       6 &       10 \\
         PyTorch 1.0 &           vgg16\_bn &  13.59 &   71.15 &  77.87 &   97.41 &      19 &       12 &      17 &       17 \\
         Keras 2.2.4 &        densenet169 &  13.94 &   70.87 &  78.85 &   98.18 &      17 &       13 &      15 &        5 \\
         PyTorch 1.0 &        densenet161 &  15.13 &   70.73 &  80.11 &   98.04 &       7 &       14 &       8 &        7 \\
         Keras 2.2.4 &        densenet121 &  13.94 &   70.59 &  76.40 &   97.48 &      18 &       15 &      20 &       16 \\
         PyTorch 1.0 &          resnet152 &  15.27 &   70.45 &  81.79 &   97.83 &       5 &       16 &       4 &       12 \\
         PyTorch 1.0 &              vgg11 &  12.96 &   70.38 &  75.49 &   97.27 &      25 &       17 &      27 &       18 \\
         PyTorch 1.0 &           vgg13\_bn &  12.68 &   69.89 &  75.84 &   96.99 &      27 &       18 &      25 &       20 \\
         PyTorch 1.0 &              vgg13 &  13.03 &   69.47 &  76.40 &   96.78 &      24 &       19 &      21 &       24 \\
         Keras 2.2.4 &       nasnetmobile &  14.15 &   69.40 &  79.27 &   96.85 &      15 &       20 &      14 &       21 \\
         Keras 2.2.4 &        densenet201 &  15.20 &   69.19 &  80.11 &   97.76 &       6 &       21 &       9 &       13 \\
         Keras 2.2.4 &        mobilenetV2 &  14.57 &   68.63 &  75.84 &   96.57 &      12 &       22 &      24 &       26 \\
         Keras 2.2.4 &  inceptionresnetv2 &  17.23 &   68.42 &  83.40 &   96.85 &       3 &       23 &       2 &       22 \\
         Keras 2.2.4 &           xception &  17.65 &   68.28 &  82.07 &   97.62 &       2 &       24 &       3 &       14 \\
         Keras 2.2.4 &        inceptionv3 &  16.11 &   68.28 &  80.25 &   96.78 &       4 &       25 &       7 &       23 \\
         Keras 2.2.4 &              vgg19 &  11.83 &   68.07 &  73.95 &   95.52 &      29 &       26 &      29 &       30 \\
         Keras 2.2.4 &          mobilenet &  14.36 &   67.58 &  73.60 &   96.08 &      13 &       27 &      31 &       27 \\
         Keras 2.2.4 &           resnet50 &  14.85 &   66.81 &  76.12 &   95.73 &       9 &       28 &      23 &       28 \\
         Keras 2.2.4 &        nasnetlarge &  19.61 &   66.32 &  84.24 &   96.57 &       1 &       29 &       1 &       25 \\
         Keras 2.2.4 &              vgg16 &  12.82 &   66.11 &  74.09 &   95.66 &      26 &       30 &      28 &       29 \\
         PyTorch 1.0 &       inception\_v3 &  14.92 &   65.62 &  75.56 &   95.38 &       8 &       31 &      26 &       31 \\
         PyTorch 1.0 &      squeezenet1\_0 &   9.66 &   63.66 &  60.50 &   91.88 &      32 &       32 &      34 &       33 \\
         PyTorch 1.0 &      squeezenet1\_1 &   9.38 &   62.54 &  61.97 &   92.30 &      33 &       33 &      33 &       32 \\
         PyTorch 1.0 &            alexnet &  11.06 &   58.96 &  62.61 &   89.29 &      31 &       34 &      32 &       34 \\
        \bottomrule
        \end{tabular}
    }
\vskip -0.1in
\end{table*}

The dramatic changes in ranking shown in Table \ref{tab:imagenet_benchmarking_table} may be explained by overfitting to the validation set when these models are trained, which can occur inadvertently during hyper-parameter tuning, or by overfitting to the noise in the training set. These results also suggest that keeping some correct labels on a secret correctable set of label errors may provide a useful framework for detecting overfitting on test sets toward a more reliable approach for benchmarking generalization accuracy across ML models.

The benchmarking experiment was replicated on CIFAR-10 in addition to ImageNet. The individual accuracies for CIFAR-10 are reported in Table \ref{tab:cifar10_benchmarking_table}. Similar to ImageNet, lower capacity models tend to outperform higher capacity models when benchmarked using corrected labels (instead of the original, erroneous labels).

Whereas traditional notions of benchmarking generalization accuracy assume the train and test distributions are the same, this is nonsensical in the case of noisy training data --- the test dataset should never contain noise because in real-world applications, we want a trained model to predict the error-free outputs on unseen examples, and benchmarking should measure as such. In two independent experiments in ImageNet and CIFAR-10, we observe that models, pre-trained on the original (noisy) datasets, with less expressibility (e.g., ResNet-18) tend to outperform higher capacity models (e.g., NASNet) on the corrected test set labels.

\begin{table*}[ht]
\setlength\tabcolsep{2pt} % Makes table columns tighter
    \caption{Individual CIFAR-10 accuracy scores for Sub-figure \ref{fig:cifar10_orig_vs_corrected} with \emph{agreement threshold = 3 of 5}. Acc@1 stands for the top-1 validation accuracy on the correctable set ($n = 18$) of original CIFAR-10 examples and labels. See Table \ref{tab:imagenet_benchmarking_table} caption for more details. Discretization of accuracies occurs due to the limited number of corrected examples on the CIFAR-10 test set.}
    \label{tab:cifar10_benchmarking_table}
    \center
    \resizebox{.9\linewidth}{!}{ %Completely zooms in or zooms out (shrinks) entire table!
        \begin{tabular}{llrrrrrrrr}
        \toprule
            Platform &         Model &  Acc@1 &  cAcc@1 &   Acc@5 &  cAcc@5 &  Rank@1 &  cRank@1 &  Rank@5 &  cRank@5 \\
        \midrule
         PyTorch 1.0 &     googlenet &  55.56 &   38.89 &   94.44 &   94.44 &       1 &       10 &      13 &       13 \\
         PyTorch 1.0 &      vgg19\_bn &  50.00 &   38.89 &  100.00 &  100.00 &       2 &       11 &       7 &        7 \\
         PyTorch 1.0 &   densenet169 &  44.44 &   50.00 &  100.00 &  100.00 &       5 &        4 &       2 &        2 \\
         PyTorch 1.0 &      vgg16\_bn &  44.44 &   44.44 &  100.00 &  100.00 &       3 &        8 &       5 &        5 \\
         PyTorch 1.0 &  inception\_v3 &  44.44 &   33.33 &  100.00 &  100.00 &       6 &       12 &       8 &        8 \\
         PyTorch 1.0 &      resnet18 &  44.44 &   55.56 &   94.44 &  100.00 &       4 &        2 &      10 &       10 \\
         PyTorch 1.0 &   densenet121 &  38.89 &   50.00 &  100.00 &  100.00 &       8 &        5 &       3 &        3 \\
         PyTorch 1.0 &   densenet161 &  38.89 &   50.00 &  100.00 &  100.00 &       9 &        6 &       4 &        4 \\
         PyTorch 1.0 &      resnet50 &  38.89 &   44.44 &  100.00 &  100.00 &       7 &        9 &       6 &        6 \\
         PyTorch 1.0 &  mobilenet\_v2 &  38.89 &   27.78 &  100.00 &  100.00 &      10 &       13 &       9 &        9 \\
         PyTorch 1.0 &      vgg11\_bn &  27.78 &   66.67 &  100.00 &  100.00 &      11 &        1 &       1 &        1 \\
         PyTorch 1.0 &      resnet34 &  27.78 &   55.56 &   94.44 &  100.00 &      13 &        3 &      11 &       11 \\
         PyTorch 1.0 &      vgg13\_bn &  27.78 &   50.00 &   94.44 &  100.00 &      12 &        7 &      12 &       12 \\
        \bottomrule
        \end{tabular}
    }
\vskip -0.1in
\end{table*}

\clearpage
\section{Expert label review details}
\label{sec:expertdetails}

To mitigate possible bias in our expert reviewing process, we did not show reviewers whether a particular image was CL-flagged or not, and we randomized whether a CL-flagged or non CL-flagged image was shown first for each ImageNet class. We also randomized whether the given or predicted label was the first or second choice offered to the reviewer. We did not however randomize the class order as reviewing was much more efficient when the classes were presented in order (required less drastic context switching) and helped reviewers to learn while reviewing, especially for taxonomies with many related classes (e.g., dog breeds). The three authors of this paper, aided by an experienced data labeler, served as these expert reviewers, spending around 67 seconds in total on average to review each image label (14x more time than MTurk workers) and around 109 seconds on average to review the images where a second phase was required for the expert reviewers to come to consensus due to disagreement (28x more time than MTurk workers). 

There were 66 ImageNet classes (out of the 1000) that had no CL-flagged image in the validation set. For these classes, the experts could not review a CL-flagged image, but experts still reviewed a non CL-flagged image. Thus, 1934 images were reviewed by experts (934 CL-flagged and 1000 non-CL flagged). 
These images were assigned into 3 non-disjoint evenly-sized partitions (one for each expert to review) such that each image was reviewed by at least 2 experts. Expert reviewer 1 was assigned images from classes 1-666. Expert reviewer 2 was assigned classes 1-333 and 667-1000. Expert reviewer 3 was assigned classes 334-1000. After independently reviewing the images (spending 54 seconds per image, on average), experts disagreed on 438 images. The experts subsequently discussed each of these images to reach a consensus decision (spending 55 seconds on average in discussions to come to consensus on a choice for each label). Table \ref{tab:expert_count} counts the different types of label issues identified by experts in the CL-flagged and non-CL flagged images, from which we computed the percentages reported in Table \ref{tab:expert_percent}. 

The time spent for expert review in Table \ref{tab:expert_percent} is computed as: (1934 / 1934 ) * 54 seconds + (438 / 1934) * 55 seconds = 67 seconds (i.e., time spent on average for all 1934 images for independent expert review + additional time spent on the 438 images requiring experts to discuss their choices and come to agreement).

In some cases, experts agreed that neither the given nor the predicted label was appropriate, but Mechanical Turk workers chose the predicted label. These were tricky cases which often required careful scrutiny to identify the true class of the given image. Figure~\ref{fig:mturk-mistake} shows an example of such a case, where the image clearly doesn't match the ImageNet given label, and upon close inspection, doesn't match the predicted label either.

\vspace*{5mm}
\begin{table}[!h]
\caption{Counts of various types of label issues  identified by experts in CL-flagged examples vs non-CL flagged examples from ImageNet (see Section \ref{sec:expertreview}). Here, count(errors) =
count(correctable) + count(multi-label) + count(neither) + count(non-agreement). Also, count(total) = count(non-errors) + count(errors). After independently making decisions about each label, experts were subsequently required to resolve any non-agreement by reaching a consensus via group deliberation. There were 66 ImageNet classes which did not have a CL-flagged error, thus only 934 CL-flagged examples were reviewed instead of 1000 (1 example for every class).
} \label{tab:expert_count}
\center
\resizebox{\textwidth}{!}{\begin{tabular}{lrrrrrrr}
\toprule
{} &  total &  non-errors &  errors &  correctable &  multi-label &  neither &  non-agreement \\
\midrule
CL (MTurk)      &    934 &         481 &     453 &          205 &           92 &       53 &           103 \\
CL (expert)     &    934 &         548 &     386 &          165 &          122 &       99 &             0 \\
non-CL (expert) &   1000 &         840 &     160 &           32 &           91 &       37 &             0 \\
\bottomrule
\end{tabular}}
\end{table}

\end{document}